\documentclass[review, 3p, number, sort&compress]{elsarticle}
\usepackage{enumitem}
\usepackage{xcolor}
\usepackage{cancel}
\usepackage{float}
\usepackage{caption}
\usepackage{subcaption}
\usepackage{graphicx}
\usepackage[hidelinks]{hyperref}
\usepackage{xspace}
\usepackage{mdframed}
\usepackage[linesnumbered, ruled, vlined]{algorithm2e}
\usepackage{threeparttable}
\usepackage{booktabs}  
\usepackage{tabularx}
\usepackage[group-separator={,}, group-minimum-digits=4]{siunitx}
\usepackage[pagewise]{lineno}
\usepackage[normalem]{ulem}

\usepackage{amsmath}
\usepackage{amssymb}
\usepackage{empheq}

\usepackage{cleveref}

\SetCommentSty{mycommfont}
\SetAlFnt{\footnotesize\ttfamily}
\SetAlCapFnt{\small}        
\SetAlCapNameFnt{\small}    
\SetKwSty{mykwfont}
\SetFuncSty{texttt}
\SetFuncArgSty{texttt}
\SetArgSty{texttt}
\SetProcNameSty{texttt}
\SetProcArgSty{texttt}

\newcommand{\Rnum}[1]{\mathbb{R}^{#1}}  

\renewcommand*{\vec}[1]{{\mathchoice  
    {\boldsymbol{\displaystyle{#1}}}  
    {\boldsymbol{\textstyle{#1}}}  
    {\boldsymbol{\scriptstyle{#1}}}  
    {\boldsymbol{\scriptscriptstyle{#1}}}  
}}
\newcommand*{\mat}[1]{\vec{{#1}}}  
\newcommand*{\inv}{^{\mkern-1.5mu\mathsf{-1}}}  
\newcommand*{\tran}{^{\mkern-1.5mu\mathsf{T}}}

\DeclareMathOperator*{\argmin}{arg\,min}
\newcommand*{\opt}{^{\mkern-1.5mu\mathsf{opt}}}

\newcommand*{\diff}{\mathop{}\!\mathrm{d}}  
\newcommand*{\deriv}[2]{\frac{\diff {#1}}{\diff {#2}}}  
\newcommand*{\pderiv}[2]{\frac{\partial {#1}}{\partial {#2}}}  
\newcommand*{\pderivi}[2]{\partial {#1} \fracslash \partial {#2}}  
\newcommand*{\tderiv}[2]{\frac{\diff {#1}}{\diff {#2}}}  
\newcommand*{\tderivi}[2]{{\diff {#1}} \fracslash {\diff {#2}}}  

\newcommand{\fracslash}{\,/\,}
\newcommand{\rvar}{x}  
\newcommand{\rvec}{\vec{\rvar}}  
\newcommand{\nrvars}{N}  
\newcommand{\param}{\alpha}  
\newcommand{\parvec}{\vec{\param}}  
\newcommand{\npars}{P}  
\newcommand{\density}{f}  
\newcommand{\cdf}{F}  
\newcommand{\pseudocdf}{\widehat{\cdf}}  
\newcommand{\condon}{\,\big|\,}  
\newcommand{\support}[1]{\Omega_{#1}}  
\newcommand{\domain}[1]{\Omega_{#1}}  
\DeclareMathOperator{\erf}{erf}  

\newcommand{\DistroSA}{\texttt{DistroSA}\xspace}
\newcommand{\OneDAlg}{\texttt{1D Alg}\xspace}
\newcommand{\FullInv}{\texttt{Full Inv}\xspace}
\newcommand{\DiagApprox}{\texttt{Diag Approx}\xspace}
\newcommand{\InterpFull}{\texttt{Interp Full}\xspace}
\newcommand{\InterpDiag}{\texttt{Interp Diag}\xspace}

\SetKwProg{Fn}{Function}{:}{}
\SetKwFunction{swap}{swap}
\SetKwFunction{interp}{interpolate}
\SetKwFunction{interpnd}{interpolate}
\SetKwFunction{condprob}{cond\_1d}
\SetKwFunction{linsolve}{solve}
\SetKwFunction{getconditionals}{get\_conditionals}
\SetKwFunction{getcdfoned}{get\_cdf\_1d}
\newcommand{\getcdfonedText}{\texttt{get\_cdf\_1d}\xspace}
\newcommand{\getconditionalsText}{\texttt{get\_conditionals}\xspace}
\newcommand{\getsensitivityonedText}{\texttt{get\_sensitivity\_1d}\xspace}
\DontPrintSemicolon

\begin{document}

  \begin{frontmatter}
    \title{Distributional Sensitivity Analysis: Enabling Differentiability in Sample-Based Inference}

    \author{Pi-Yueh Chuang\corref{cor}}
    \ead{pchuang@anl.gov}
    \cortext[cor]{Corresponding author}
    \author{Ahmed Attia}
    \ead{aattia@anl.gov}
    \author{Emil Constantinescu}
    \ead{emconsta@anl.gov}
    \affiliation{
      organization={%
        Mathematics and Computer Science Division, %
        Argonne National Laboratory%
      },
      city={Lemont},
      state={IL 60439},
      country={USA}
    }

\begin{abstract}
  This work introduces a mathematical framework for estimating the space–parameter sensitivity of random samples in arbitrary dimensions.
  Such sensitivity effectively acts as gradients of random samples with respect to distributional parameters, which are essential in sample-based inverse problems in nuclear physics, such as inferring quantum correlation functions.
  We present two analytical formulae for sensitivity and gradient estimation.
  The first interprets sensitivity as the partial derivatives of the inverse mapping of 1-D conditional distributions.
  The second, suited for optimization methods that tolerate inexact gradients, applies a diagonal approximation that reduces computational cost with minimal accuracy loss.
  When closed forms are unavailable, four second-order numerical algorithms are provided to approximate both expressions.
  Verification and validation studies confirm the correctness of these algorithms and the effectiveness of the proposed formulae.
  A nuclear physics application demonstrates how the framework enables uncertainty quantification and parameter inference for quantum correlation functions.
  Unlike existing approaches, our method requires neither model fitting nor knowledge of sampling algorithms or high-dimensional integrals, making it suitable for black-box or simulation-based samplers.
  Moreover, it renders arbitrary sampling subroutines \emph{differentiable}, facilitating integration into deep learning and automatic differentiation frameworks.
  Algorithmic details and open-source implementations are provided to ensure reproducibility and promote further development.
\end{abstract}

  \end{frontmatter}



\section{Introduction}
\label{sec:intro}


Quantifying the sensitivities of random samples drawn from a parametric probability distribution with respect to its parameters plays a key role in sample-based inverse problems.
Such inverse problems are common in nuclear physics applications, including the inference of quantum correlation functions~\cite{cranmer_frontier_2020, braga_toward_2025}.
In these applications, the objective is to infer the parameters of a probabilistic model that best describes the training data, such as experimental observations from particle colliders.
This is achieved by defining a suitable distance measure between the training data and random samples generated from the probabilistic model, and subsequently minimizing this distance with respect to the model parameters.
Common stochastic loss functions used to quantify this distance include the Kullback--Leibler (KL) divergence and the energy score~\cite{constantinescu_statistical_2020, gneiting_strictly_2007}.
The sensitivities of random samples thus act as effective gradients or Jacobians, guiding the minimization of these stochastic loss functions~\cite{pflug_sampling_1989, mohamed_monte_2020}.

In particular, the gradients of random samples naturally appear in the chain-rule decomposition of the loss gradient with respect to the distributional parameters---whether this decomposition is carried out manually through direct calculations or implicitly via automatic differentiation.
For a loss function $L(\rvec_1, \rvec_2, \ldots)$, its gradient with respect to the parameters $\parvec$ can be expressed as $\nabla_{\parvec} L = \sum_i \left( \nabla_{\parvec} \rvec_i \right)^\mathsf{T} \nabla_{\rvec_i} L$, where $\rvec_i$ denotes the $i$-th realization of the random variable $\rvec$ following the probabilistic model parameterized by $\parvec$.
Here, $\nabla_{\parvec}\rvec_i \equiv \left.\nabla_{\parvec}\rvec\right|_{\rvec=\rvec_i}$ represents the gradient of the $i$-th realization with respect to the parameters.

A challenge, however, arises when calculating $\nabla_{\parvec}\rvec_i$.
Specifically, when $\parvec$ changes, we \emph{observe new realizations} from the updated distribution rather than \emph{move existing realizations} to new locations in the sample space.
As a result, there is no direct correspondence between realizations before and after $\parvec$ changes.
In other words, the symbol $\rvec_i$ before and after a change in $\parvec$ does not represent the same object.
This nature introduces discontinuities that invalidate the classical limit-based (Cauchy) definition of a derivative unless additional assumptions or reinterpretations are introduced.

\subsection{Existing Methods}
\label{subsec:state_of_the_art}

Existing methods addressing this challenge can be broadly divided into two families.
The first resolves the ambiguity in differentiating random samples by reformulating or approximating the loss function.
A representative example is the distributional derivative~\cite{fu_chapter_2006}, which exploits the fact that many stochastic losses---such as the KL divergence and energy score---can be expressed as expectations $\mathbb{E}_{\rvec\sim\density(\rvec; \parvec)}[l(\rvec)]$, where $l(\rvec)$ is the loss kernel and $\density$ the parameterized probability density function (PDF).
Writing this expectation as an integral, $\int l(\rvec) f(\rvec; \parvec)\diff \rvec$, allows differentiation with respect to $\parvec$ as $\int l(\rvec)\nabla_{\parvec}f(\rvec; \parvec)\diff \rvec$, eliminating the need to generate realizations $\rvec_i$ or compute $\nabla_{\parvec}\rvec_i$.
However, this approach cannot be applied when $\density(\rvec; \parvec)$ is non-differentiable.
Moreover, the integral still requires numerical evaluation, which becomes intractable in high dimensions for grid-based methods (e.g., trapezoidal or Gaussian quadrature).
Monte Carlo integration~\cite{kleijnen_optimization_1996, wingate_automated_2013} is typically used instead to approximate $\int l(\rvec)\nabla_{\parvec}f(\rvec; \parvec)\diff \rvec$.
Although it mitigates the curse of dimensionality, it converges slowly and requires large sample sizes.
In \Cref{subsec:validation}, we use this approach as a baseline for comparison with our proposed method.

Surrogating the mapping from parameters directly to the loss, which allows gradients to be obtained by differentiating the surrogate, is another common strategy in this family.
Earlier studies favored simple surrogate forms to ease training and enable analytical differentiation~\cite{box_experimental_1951, spall_introduction_2003, queipo_surrogate-based_2005}, whereas recent work employs deep neural networks and adversarial training to exploit the advances in high-performance computing and automatic differentiation~\cite{schulman_gradient_2015, louppe_adversarial_2020}.
Although this approach avoids high-dimensional integration and accommodates non-differentiable PDFs, the fidelity of surrogate models can introduce additional errors that degrade overall performance.

The second family, known as pathwise derivatives~\cite{fu_chapter_2006}, defines or interprets gradients of random samples with respect to parameters by establishing deterministic and differentiable mappings between them.
The most common approach is the reparameterization trick~\cite{naesseth_reparameterization_2020}, which expresses a random variable as a deterministic function of the parameters and auxiliary inputs.
Gradients are then computed by differentiating this function with respect to the parameters while treating the auxiliary inputs as constants.
For instance, a multivariate Gaussian realization can be represented as $\rvec=\vec{\mu} + L(\vec{\nu})\Phi^{-1}(\vec{u})$, where $\vec{\mu}$ is the mean, $L(\vec{\nu})$ the lower Cholesky factor of the covariance matrix $\vec{\nu}$, $\Phi^{-1}$ the standard normal quantile function, and $\vec{u}$ a realization of a uniform random vector.
By fixing $\vec{u}$ and differentiating this mapping, one obtains $\nabla_{\vec{\mu}, \vec{\nu}} \rvec$ for Gaussian samples.
Note that Gaussian samples need not be generated using this transformation.
It serves only as a means for gradient estimation.

A major challenge of the reparameterization trick is constructing such transformations for complex or black-box distributions.
Additionally, when samples are not generated through these transformations (for instance, when obtained via general-purpose samplers), the corresponding auxiliary inputs must first be reconstructed (e.g., recovering $\vec{u}$ in the Gaussian case) before differentiation.
Recent research leverages deep learning to approximate or surrogate the mapping and inverse mapping between distributional parameters and realizations, thereby enabling reparameterization for arbitrary distributions in principle.
Representative methods include normalizing flows~\cite{papamakarios_normalizing_2021, du_unifying_2025}, variational autoencoders~\cite{kingma_auto-encoding_2022}, and physics-informed neural networks~\cite{li_surrogate_2023}.
Nonetheless, their practical effectiveness---in terms of computational cost and accuracy---remains an open question.

\subsection{Problem Definition}
\label{subsec:prob_des}

This study focuses on the numerical computation of \emph{space--parameter sensitivity} at realizations of a random vector, defined as
\begin{equation}
  \Big[\nabla_{\param} \rvec\Big]_{
    \substack{
      \rvec \in \support{\rvec} \\
      \parvec \in \support{\parvec}
    }
  }
  =
  \begin{bmatrix}
    \pderiv{\rvar_{1}}{\param_{1}} & \pderiv{\rvar_{1}}{\param_{2}}
      & \cdots & \pderiv{\rvar_{1}}{\param_{\npars}} \\
    \pderiv{\rvar_{2}}{\param_{1}} & \pderiv{\rvar_{2}}{\param_{2}}
      & \cdots & \pderiv{\rvar_{2}}{\param_{\npars}}  \\
    \vdots & \vdots & \ddots & \vdots \\
    \pderiv{\rvar_{\nrvars}}{\param_{1}} & \pderiv{\rvar_{\nrvars}}{\param_{2}}
      & \cdots & \pderiv{\rvar_{\nrvars}}{\param_{\npars}}
  \end{bmatrix}_{
    \substack{
      \rvec \in \support{\rvec} \\
      \parvec \in \support{\parvec}
    }
  }
  \quad\,;\,\,\rvec \in \support{\rvec}
    \subseteq \domain{\density} \subseteq \Rnum{\nrvars} \,,
  \label{eqn:space_parameter_sensitivity}
\end{equation}
where $\rvec \coloneqq \left[\rvar_1, \rvar_2, \ldots, \rvar_\nrvars\right]\tran$ is a realization of an $N$-D random vector with normalized PDF $\density(\rvec;\,\parvec)$.
Here $\domain{\density}$ is the domain of $\density$, $\support{\rvec} \equiv \left\{\rvec \condon \density (\rvec;\;\parvec) > 0 \right\}$ is its support, and $\parvec \in \support{\parvec} \subseteq \Rnum{\npars}$ is the vector of distributional parameters.
This sensitivity should be independent of the sampling procedure used to produce realizations.

The problem here is that equation \eqref{eqn:space_parameter_sensitivity} is ill-defined.
Measure--theoretic differentiation does not apply to a random variable or its realizations.
Cauchy differentiation also fails, because realizations before and after perturbing $\parvec$ lack a canonical correspondence.
Hence additional context or interpretations are needed to define \eqref{eqn:space_parameter_sensitivity}.

As an example, for a univariate continuous distribution, one may define the sensitivity as the rate of change of samples at a fixed cumulative distribution function (CDF) level with respect to parameter perturbations.
In this context (i.e., fixed CDF level), \eqref{eqn:space_parameter_sensitivity} is equivalent to evaluating the partial derivative of the inverse CDF.
As another example, for a multivariate Gaussian (as shown in \Cref{subsec:state_of_the_art}), one may define the sensitivity as the partial derivatives of $\rvec \equiv \rvec(\vec{u};\,\vec{\mu}, \vec{\nu}) = \vec{\mu} + L(\vec{\nu})\Phi^{-1}(\vec{u})$, holding the auxiliary vector $\vec{u}$ fixed as parameters ($\vec{\mu}$ and $\vec{\nu}$) vary.
Unfortunately, these approaches do not easily generalize to arbitrary distributions and dimensionalities.

This work develops a mathematical framework that supplies the needed context to make \eqref{eqn:space_parameter_sensitivity} well-defined and computable for general parametric families, even when only an unnormalized density is available.
It also provides robust, practical numerical algorithms to evaluate \eqref{eqn:space_parameter_sensitivity} for high--dimensional and black-box distributions under that framework.
Attaining this objective is critical for applications including Bayesian modeling and distribution fitting in high--dimensional Bayesian inverse problems, as well as machine learning models that rely on automatic differentiation when the forward pass involves random sampling.

Note that the sensitivity in \eqref{eqn:space_parameter_sensitivity} concerns only the realizations (samples).
For notational simplicity, we drop subscripts hereafter and write $\nabla_{\parvec} \rvec$ with $\rvec \in \support{\rvec}$ and $\parvec \in \support{\parvec}$ understood.
Additionally, \eqref{eqn:space_parameter_sensitivity} has the same form as the Jacobian of $\rvec$ with respect to $\parvec$ and is used as such for gradient--based minimization methods.
We therefore use the terms sensitivity, Jacobian, and gradient interchangeably.

\begin{mdframed}
    In summary, our work focuses on evaluating \eqref{eqn:space_parameter_sensitivity} under the following assumptions:
    \begin{enumerate}
      \item The PDF is non-degenerate and positive $\density(\rvec;\,\parvec) > 0$ on a connected domain $\support{\rvec}$, and this support is independent of $\parvec$.
      \item $\density(\rvec;\,\parvec)$ is a smooth function in $\rvec$ and is differentiable in $\parvec$; that is, $\pderivi{\density(\rvec;\,\parvec)}{\parvec}$ exists.
    \end{enumerate}
\end{mdframed}

While our work and test cases focus primarily on distributions satisfying these assumptions, the method can remain applicable beyond them.
For example, \Cref{subsec:when_f_ge_zero} discusses the case where $\Omega_{\rvar}$ comprises multiple disjoint domains.

\subsection{Our Contributions}

We introduce two analytical formulae (see \Cref{sec:methods}) to interpret and compute the sensitivity~\eqref{eqn:space_parameter_sensitivity}.
These formulae differ in their levels of mathematical approximation.
We also provide four numerical algorithms to evaluate these formulae for multivariate distributions.
All four algorithms achieve second-order error convergence but have different numerical strategies, leading to distinct computational time, peak memory, and accuracy.
Additionally, we present a simplified formula and algorithm for 1-D distributions, a special case that enjoys substantial analytical and algorithmic simplification.
Implementations are available in our open-source package \DistroSA\ (Distributional Sensitivity Analysis)~\cite{attia_distrosa_2025, attia_distrosa_2025-1}.

A central feature of our method is independence from the sampling procedure, which makes it suitable as a backward-propagation mechanism for arbitrary computer sampling subroutines and a useful complement to automatic differentiation frameworks such as PyTorch.

Our first formula, \eqref{eqn:nd_sensitivity_def}, bears similarity to the approach by Figurnov et al.~\cite{figurnov_implicit_2018} but requires only 1-D conditionals in multivariate settings, whereas their approach invokes the full Rosenblatt mapping and thus high-dimensional integration.
This difference yields a computational advantage for multivariate distributions.
Our second formula, \eqref{eqn:nd_sensitivity_approx}, to the best of our knowledge, is novel.
The term \emph{implicit reparameterization}, as introduced by Figurnov et al., still aptly describes our approach: we assume such a reparameterization exists but do not require its explicit construction, and this assumption alone suffices to derive the core idea.

The remainder of the paper is organized as follows.
\Cref{sec:methods} derives the method.
\Cref{sec:benchmarks} provides verification and validation (V\&V) and demonstrates a nuclear-physics application.
\Cref{sec:discussion} concludes with potential improvements and future directions.


\section{Method: Distributional Space-Parameter Sensitivity Analysis}
\label{sec:methods}

This section presents our method for computing the distributional space–parameter sensitivity in \eqref{eqn:space_parameter_sensitivity}.
We first develop the 1-D case in \Cref{subsec:method_1d}, extend to multivariate distributions in \Cref{subsec:method_nd}, and introduce a diagonal approximation in \Cref{subsec:inverse_mat_approx}.
Numerical algorithms are provided as well.
We discuss applicability to PDFs with zero-density regions in \Cref{subsec:when_f_ge_zero}.
\Cref{subsec:comp_cost} analyzes the computational costs and practical considerations of the presented numerical algorithms.

For clarity, we assume throughout the derivations that the density $\density$ is normalized.
This simplifies the mathematics but is not required in practice: the algorithms use numerical integration that automatically handles unnormalized densities.

\subsection{1-D Probability Distributions}
\label{subsec:method_1d}

We focus on 1-D distributions, first for scalar parameters (\Cref{subsubsec:1d_scalar_param}) and then for vector-valued parameters (\Cref{subsubsec:1d_vector_param}).

\subsubsection{Scalar Parameter.}
\label{subsubsec:1d_scalar_param}

Consider a 1-D random variable $\rvar\in \support{\rvec}\subseteq\mathbb{R}$ with PDF $\density(\rvar; \param)$ parameterized by $\param\in\support{\parvec}\subseteq \mathbb{R}$.
In this case, the distributional space-parameter sensitivity \eqref{eqn:space_parameter_sensitivity} simplifies to
\begin{equation}
  \nabla_{\param} \rvar
  =
  \pderiv{\rvar}{\param}
  \,.
  \label{eqn:1d_1p_sensitivity}
\end{equation}
We emphasize the use of the partial derivative ($\partial$) rather than the total derivative ($\diff$) in \eqref{eqn:1d_1p_sensitivity}.
Since the parameter is non-random, a realization of $\rvar$ cannot be expressed solely as a function of the parameter.
Other independent variables must exist and typically contribute to the randomness of $\rvar$ and to the generation of its realizations.
These additional independent variables, whether known or unknown, must be excluded from the effect of $\param$ on $\rvar$.

When the PDF is smooth and strictly positive, the CDF $\cdf$ is strictly monotone and bijective, so the inverse CDF (quantile) $\cdf\inv$ exists:
\begin{subequations}
  \begin{empheq}[left=\empheqlbrace]{align}
    u &=
    \cdf(x;\,\param)
    \equiv
    \int_{-\infty}^x \density(z;\,\param) \diff z \,,
    \label{eqn:1d_cdf}
    \\
    x &= \cdf\inv(u,\,\param) \,,
    \label{eqn:1d_inv_cdf}
  \end{empheq}
\end{subequations}
where $u \in [0, 1]$ is the cumulative probability from $-\infty$ up to $x$.
\Cref{eqn:1d_inv_cdf} yields the inverse transform sampling~\cite{robert_monte_2004}: draw $u\sim\mathcal{U}(0,1)$ and set $x=\cdf\inv(u,\param)$.
In this sampling scheme, $u$ is independent of $\param$ and does not contribute to $\pderivi{\rvar}{\param}$.
Thus, for sensitivity analysis we may, without loss of generality, treat any realization as if it were obtained via \eqref{eqn:1d_inv_cdf} and hold $u$ fixed.
Then $\pderivi{\rvar}{\param}=\pderivi{\cdf\inv}{\param}$, the simplest form of the reparameterization trick.

Computing $\pderivi{\cdf\inv}{\param}$ directly requires access to $\cdf\inv$ and the associated $u$ for each realization, which can be costly or infeasible, especially beyond 1-D.
Instead, we use \eqref{eqn:1d_cdf} and note that $u$ and $\param$ are the only independent variables on which $\rvar$ depends in \eqref{eqn:1d_inv_cdf}.
The total differential is
\begin{equation*}
  \diff \rvar
  =
  \pderiv{\rvar}{\param} \diff \param
  +
  \pderiv{\rvar}{u} \diff u
  \,.
\end{equation*}
Therefore,
\begin{equation}\label{eqn:tot_deriv_1d}
  \nabla_{\param} \rvar
  =
  \pderiv{\rvar}{\param}
  =
  \left.\tderiv{\rvar}{\param}\right|_{\diff u = 0}
  \,.
\end{equation}
To enforce $\diff u = 0$, apply the total differential to~\eqref{eqn:1d_cdf}:
\begin{equation}
  \diff u  \stackrel{\eqref{eqn:1d_cdf}}{=} 0
  \quad \Rightarrow \quad
  \pderiv{\cdf}{\rvar} \diff \rvar
  +
  \pderiv{\cdf}{\param} \diff \param
  =
  0
  \quad \Rightarrow \quad
  \deriv{\rvar}{\param}
  = -
  \left(1 \middle/ \pderiv{\cdf}{\rvar} \right)
  \pderiv{\cdf}{\param}
  =
  -
  \frac{1}{\density}
  \pderiv{\cdf}{\param}
  \,,
  \label{eqn:1d_1p_tmp_1}
\end{equation}
and we arrive at
\begin{equation}
  \nabla_{\param} \rvar
  = \pderiv{\rvar}{\param}
  \stackrel{(\ref{eqn:tot_deriv_1d}, \ref{eqn:1d_1p_tmp_1})}{=}
  -
  \frac{1}{\density}
  \pderiv{\cdf}{\param}
  \,.
  \label{eqn:1d_1p_sensitivity_def}
\end{equation}

\Cref{eqn:1d_1p_sensitivity_def} defines the space-parameter sensitivity for a 1-D distribution with a single scalar parameter $\param$.
It does not require knowledge of how a realization of $\rvar$ is generated or the exact value of $u$ associated with this realization.
More importantly, it does not require $\cdf\inv$ to be explicitly known or computable.
This property will be crucial in extending the method to multivariate distributions.

\subsubsection{Vector-Valued Parameter.}
\label{subsubsec:1d_vector_param}

The extension of \eqref{eqn:1d_1p_sensitivity_def} to a vector-valued parameter is straightforward.
By replacing the scalar parameter in \eqref{eqn:1d_1p_sensitivity_def} with a parameter vector $\parvec:=\left[\param_1, \ldots, \param_{\npars}\right]\tran$, we obtain:
\begin{equation}
    \nabla_{\parvec} \rvar
    =
    \pderiv{\rvar}{\parvec}
    =
    \begin{bmatrix}
      \dfrac{\partial\rvar}{\partial\param_{1}},
        & \ldots,
        & \dfrac{\partial\rvar}{\partial\param_{\npars}}
    \end{bmatrix}\tran
    =
    - \frac{1}{\density}
    \pderiv{\cdf}{\parvec}
    =
    - \frac{1}{\density}
    \begin{bmatrix}
      \dfrac{\partial\cdf}{\partial\param_{1}},
        & \ldots,
        & \dfrac{\partial\cdf}{\partial\param_{\npars}}
    \end{bmatrix}\tran \,.
  \label{eqn:1d_sensitivity_def}
\end{equation}
A closed form for \eqref{eqn:1d_sensitivity_def} is rare because it requires $\pderivi{\cdf}{\parvec}$.
In practice, this is challenging, particularly when $\density$ is evaluated via black-box simulation.
Therefore, we usually rely on numerical approximations.

\Cref{alg:get_sensitivity_1d} describes the solution algorithm that evaluates \eqref{eqn:1d_sensitivity_def} at a single spatial point.
\Cref{tab:symbols} lists the common symbols used in this and all later algorithms.
The algorithm uses second-order central differences to approximate the derivatives in \eqref{eqn:1d_sensitivity_def} for robustness and simplicity with arbitrary PDFs.
Interpolation (Line~\ref{alg:get_sensitivity_1d:interp}) approximates the derivative at $\rvar$.
Our work uses piecewise linear interpolation to preserve overall second-order accuracy.
This algorithm is later used in our benchmark studies in \Cref{sec:benchmarks} and is labeled as \OneDAlg.

\begin{table}[htbp!]
  \centering
  \caption{Symbols used in algorithms.}
  \label{tab:symbols}
  \small
  \renewcommand{\arraystretch}{1.2}
  \begin{tabularx}{\textwidth}{cX}
    \toprule
      Symbol & Description \\
    \midrule
      $\nrvars$ &
        Number of spatial dimensions
    \\
      $\npars$ &
        Number of parameters
    \\
      $\density$ &
        Either a 1-D PDF ($\density\colon\Rnum{}\times\Rnum{\npars}\to\Rnum{+}$) or an N-D joint PDF ($\density\colon\Rnum{\nrvars}\times\Rnum{\npars}\to\Rnum{+}$)
    \\
      $\vec{v}$ &
        Gridline expressed as an increasing 1-D sequence of vertices ($\vec{v}=(v^1, \ldots, v^K)\in\Rnum{K}$ where $K$ is the number of vertices)
    \\
      $\vec{V}$ &
        Rectilinear grid in N-D expressed as a sequence of gridlines ($\vec{V} = (\vec{v}_1, \ldots, \vec{v}_\nrvars)$ and $\vec{v}_i = (v_i^1, \ldots, v_i^{K_i}) \in \Rnum{K_i}$, where $K_i$ is the number of vertices in the
        $i$-th spatial direction)
    \\
      $\parvec$ &
        Parameter vector ($\parvec = (\param_1, \ldots, \param_\npars) \in \Rnum{\npars}$)
    \\
      $\rvar$ &
        Space point in 1-D where the sensitivity will be estimated ($\rvar \in \support{\rvar} \subseteq \Rnum{}$)
    \\
      $\rvec$ &
        Space point in N-D where the sensitivity will be estimated ($\rvec = (x_1, \ldots, x_\nrvars) \in \support{\rvec}$)
    \\
      $\mat{X}$ &
        $M$ space points in N-D where the sensitivity will be estimated ($\mat{X} \in \Rnum{M \times \nrvars}$)
    \\
      $\epsilon$ &
        Scalar finite-differences step size ($\epsilon \in \Rnum{+}$)
    \\
    \bottomrule
   \end{tabularx}
 \end{table}

\begin{algorithm}[htbp!]
  \SetKwFunction{gradonedim}{get\_sensitivity\_1d}
  \caption{%
    1-D sensitivity evaluation at a single point. %
    This algorithm is labeled as \OneDAlg later in \Cref{sec:benchmarks}. %
    The external function \getcdfonedText is described in \Cref{alg:get_cdf_1d} in \ref{sec:algorithms}.}
  \label{alg:get_sensitivity_1d}

  \KwIn{1-D $\density$, $\vec{v}$, $\parvec$, $\rvar$, $\epsilon$}

  \KwOut{$\vec{g}$ ($\vec{g} \coloneqq \nabla_{\parvec} \rvar$ as in \cref{eqn:1d_sensitivity_def})}

  \BlankLine
  \Fn(){\gradonedim{$\density$, $\vec{v}$, $\parvec$, $\rvar$, $\epsilon$}}{

    \BlankLine
    $\vec{g} \gets \mathbf{0}_{\npars}$\;

    $c \gets$ \getcdfoned{$\density$, $\vec{v}$, $\parvec$}[2]
      \tcp*[l]{only need the normalization factor}

    $f_\rvar \gets \density(\rvar,\,\parvec) \fracslash c$
      \tcp*[l]{must be normalized}

    \BlankLine
    \tcp{run over $j$th parameter component}
    \For(){$j \gets 1$ \KwTo $\npars$}{

      $\vec{e}_j \gets$ $j$th standard basis in $\Rnum{\npars}$\;

      $\vec{\Phi}^{+} \gets$
        \getcdfoned{$\density$, $\vec{v}$, $\parvec+\epsilon\vec{e}_j$}[1]\;

      $\vec{\Phi}^{-} \gets$
        \getcdfoned{$\density$, $\vec{v}$, $\parvec-\epsilon\vec{e}_j$}[1]\;

      $\vec{\delta} \gets
        (\vec{\Phi}^{+} - \vec{\Phi}^{-}) \fracslash (2 \epsilon)$
        \tcp*[l]{element-wise subtraction \& scaling}

      $\vec{g}[j] \gets -\,$
        \interp{$\rvar$, $\vec{v}$, $\vec{\delta}$} $\fracslash\,\density_\rvar$
        \nllabel{alg:get_sensitivity_1d:interp}\;
    }

    \KwRet $\vec{g}$\;
  }
\end{algorithm}

\subsection{Multivariate Probability Distributions}
\label{subsec:method_nd}

We now extend to multivariate distributions, $\rvec \sim \density(\rvec;\,\parvec)$ with $\rvec \in \support{\rvec} \subseteq \Rnum{\nrvars}$ and $\parvec \in \support{\parvec} \subseteq \Rnum{\npars}$.
While the multivariate CDF $\cdf$ is well defined, its inverse is generally not; we therefore generalize the 1-D idea via conditional CDFs.

Let $\vec{u} \in [0,1]^N$,  with marginal components in $\mathcal{U}(0, 1)$.
We treat realizations as if generated by $(\vec{u},\,\parvec) \mapsto \rvec$ and use $\nabla_{\parvec} \rvec = \left.\tderivi{\rvec}{\parvec}\right|_{\diff\vec{u}=0}$.
We need not know $(\vec{u},\,\parvec) \mapsto \rvec$ explicitly; instead, we define $(\rvec,\,\parvec) \mapsto \vec{u}$ and enforce $\diff\vec{u}=0$.
This requires $(\rvec,\,\parvec) \mapsto \vec{u}$ to be (almost everywhere) bijective on $\support{\rvec}$ so that the inverse exists.

Let us define the mapping $\pseudocdf: (\rvec;\,\parvec) \mapsto \vec{u}$ as follows:
\begin{subequations}\label{eqn:conditional_mapping}
  \begin{align}
    \pseudocdf(\rvec;\,\parvec)
    &:=
    \left[
      F_1(\rvar_1 \condon \rvec_{-1};\,\parvec),\,
      F_2(\rvar_2 \condon \rvec_{-2};\,\parvec),\,
      \ldots,\,
      F_{\nrvars}(\rvar_{\nrvars} \condon \rvec_{-{\nrvars}};\,\parvec)
    \right]\tran
    =
    \vec{u}
    \label{eqn:nd_cdf}
    \,, \\
    F_i(\rvar_i \condon \rvec_{-i};\,\parvec)
    &=
    \int\limits_{z=-\infty}^{z=\rvar_i}
      \density_i(z \condon \rvec_{-i};\,\parvec)
      \diff z
    =
    u_i
    \label{eqn:nd_cdf_i}
    \,, \\
    \density_i(\rvar_i \condon \rvec_{-i};\,\parvec)
    &=
    \frac{\density(\rvec;\,\parvec)}{
        \int\limits_{z=-\infty}^{z=\infty}
          \density(
            \left[
              x_1,\, \ldots,\, x_{i-1},\, z,\, x_{i+1},\, \ldots,\, x_{\nrvars}
            \right]\tran
            ;\,
            \parvec
          )
          \diff z
    }
    \label{eqn:nd_pdf_i}
    \,,
  \end{align}
\end{subequations}
where $\rvec_{-i} \in \Rnum{\nrvars-1}$ denotes a vector of all elements in $\rvec \in \Rnum{\nrvars}$ except the $i$th entry $\rvar_i$.

The vector of 1-D conditional CDFs in \eqref{eqn:nd_cdf} is locally strictly increasing in each coordinate but not guaranteed globally bijective on $\support\rvec$.
In our experiments across several families, however, we did not observe non-invertibility (e.g., multivariate Gaussians fail only in degenerate cases; see \ref{sec:analytical_gaussian}).
We therefore assume the inverse exists almost everywhere and provide a remedy in \Cref{subsubsec:nd_bijectivity} if this fails.

Continuing under the assumption that the bijectivity of \eqref{eqn:nd_cdf} holds, we derive the space-parameter sensitivity by applying $\diff \vec{u} = 0$ to \eqref{eqn:nd_cdf} and get
\begin{equation}
  \diff \vec{u}
  =
  \nabla_{\rvec} \pseudocdf \diff \rvec
  +
  \nabla_{\parvec} \pseudocdf \diff \parvec
  =
  0
  \,.
\end{equation}
By noting that $\nabla_{\parvec} \rvec = \left.\frac{\diff \rvec}{\diff \parvec}\right|_{\diff \vec{u} = 0}$, this leads to
\begin{equation}
  \nabla_{\parvec} \rvec
  =
  -
  \left(\nabla_{\rvec} \pseudocdf\right)\inv \,
  \nabla_{\parvec} \pseudocdf
  =
  -
  \begin{bmatrix}
   \pderiv{\cdf_1}{\rvar_1} & \ldots & \pderiv{\cdf_1}{\rvar_{\nrvars}} \\
   \vdots & \ddots & \vdots \\
   \pderiv{\cdf_\nrvars}{\rvar_1} & \ldots & \pderiv{\cdf_\nrvars}{\rvar_\nrvars}
  \end{bmatrix}\inv
  \begin{bmatrix}
    \pderiv{\cdf_1}{\param_1} & \ldots & \pderiv{\cdf_1}{\param_{\npars}} \\
    \vdots & \ddots & \vdots \\
    \pderiv{\cdf_\nrvars}{\param_1} & \ldots & \pderiv{\cdf_\nrvars}{\param_\npars}
  \end{bmatrix} \,.
  \label{eqn:nd_sensitivity_def}
\end{equation}
\Cref{eqn:nd_sensitivity_def} gives the multivariate space–parameter sensitivity and reduces to \eqref{eqn:1d_sensitivity_def} when $\nrvars=1$.
In practice, entries of $\nabla_{\rvec} \pseudocdf$ and $\nabla_{\parvec} \pseudocdf$ are typically computed numerically.

We present two solution algorithms that numerically evaluate \eqref{eqn:nd_sensitivity_def}: \Cref{alg:get_sensitivity_nd} (labeled as \FullInv in \Cref{sec:benchmarks}) and \Cref{alg:get_sensitivity_nd_grid} (labeled as \InterpFull).
The latter is expected to be computationally cheaper but less accurate than the former for low-dimensional distributions, as shown in the cost analysis in \Cref{subsec:comp_cost}.
We describe their implementation details and key differences below.

\Cref{alg:get_sensitivity_nd} computes the sensitivity at a single spatial point by forming $\nabla_{\rvec}\pseudocdf$ and $\nabla_{\parvec}\pseudocdf$ and solving the resulting linear system there.
Cost scales linearly with the number of queried points.
We use central finite differences and piecewise multilinear interpolation~\cite{weiser_note_1988} to retain second-order accuracy and reuse the 1-D CDF calculator (\Cref{alg:get_cdf_1d}).

\begin{algorithm}[htbp!]
  \SetKwFunction{gradnd}{get\_sensitivity\_nd}
  \caption{%
    The first numerical algorithm for evaluating \cref{eqn:nd_sensitivity_def}. %
    It is labeled as \FullInv later in \Cref{sec:benchmarks}. %
    The external function \getcdfonedText is described in \Cref{alg:get_cdf_1d} in \ref{sec:algorithms}.%
  }
  \label{alg:get_sensitivity_nd}

  \KwIn{$\density$, $\vec{V}$, $\parvec$, $\rvec$, $\epsilon$}

  \KwOut{%
    $\mat{J}$ ($\mat{J} \coloneqq \nabla_{\parvec} \rvec$ as in
    \cref{eqn:nd_sensitivity_def})%
  }

  \BlankLine
  \Fn{\gradnd{$\density$, $\vec{V}$, $\parvec$, $\rvec$, $\epsilon$}}{

    \BlankLine

    $\mat{H} \gets \mathbf{0}_{\nrvars \times \nrvars}$
      \tcp*[l]{$\mat{H}$ will hold $-\nabla_{\rvec}\pseudocdf$}

    $\mat{G} \gets \mathbf{0}_{\nrvars \times \npars}$
      \tcp*[l]{$\mat{G}$ will hold $\nabla_{\parvec}\pseudocdf$}

    \BlankLine
    \tcp{constructing matrices}
    \For(\tcp*[h]{$i$-th conditional}){$i \gets 1$ \KwTo $\nrvars$}{

      \BlankLine

      $\vec{e}_i \gets$ $i$-th standard basis in $\Rnum{\nrvars}$\;

      $\density_i(s,\,\parvec) \gets \density(\rvec+(s-\rvar_i)\vec{e}_i,\,\parvec)$
        \tcp*[l]{a lambda/local function}

      \BlankLine
      \For(\tcp*[h]{$j$-th axis in space}){$j \gets 1$ \KwTo $\nrvars$}{

        \BlankLine

        \If(\tcp*[h]{conditional PDF in diagonal}){$i = j$}{
          $c \gets$ \getcdfoned{$\density_i$, $\vec{v}_i$, $\parvec$}[2]
            \tcp*[l]{the normalization factor}

          $\mat{H}[i, j] \gets - \density_i(\rvar_i,\,\parvec) \fracslash c$\;

          continue\;
        }

        \BlankLine

        $\vec{e}_j \gets$ $j$-th standard basis in $\Rnum{\nrvars}$\;

        $\density_i^{+}(s,\,\parvec) \gets
          \density(\rvec+\epsilon\vec{e}_j+(s-\rvar_i)\vec{e}_i,\,\parvec)$
          \tcp*[l]{a lambda/local function}

        $\density_i^{-}(s,\,\parvec) \gets
          \density(\rvec-\epsilon\vec{e}_j+(s-\rvar_i)\vec{e}_i,\,\parvec)$
          \tcp*[l]{a lambda/local function}

        $\vec{\Phi}^{+} \gets$
          \getcdfoned{$\density_i^{+}$, $\vec{v}_i$, $\parvec$}[1]\;
        $\vec{\Phi}^{-} \gets$
          \getcdfoned{$\density_i^{-}$, $\vec{v}_i$, $\parvec$}[1]\;

        $\vec{\delta} \gets (\vec{\Phi}^{+}-\vec{\Phi}^{-})\fracslash(2\epsilon)$\;

        $\mat{H}[i, j] \gets$ - \interp{$\rvar_i$, $\vec{v}_i$, $\vec{\delta}$}
          \tcp*[l]{remember the negative sign}
      }

      \BlankLine

      \For(\tcp*[h]{$j$-th component in parameter}){$j \gets 1$ \KwTo $\npars$}{

        $\vec{e}_j \gets$ $j$-th standard basis in $\Rnum{\npars}$\;

        $\vec{\Phi}^{+} \gets$
          \getcdfoned{$\density_i$, $\vec{v}_i$, $\parvec+\epsilon\vec{e}_j$}[1]\;

        $\vec{\Phi}^{-} \gets$
          \getcdfoned{$\density_i$, $\vec{v}_i$, $\parvec-\epsilon\vec{e}_j$}[1]\;

        $\vec{\delta} \gets (\vec{\Phi}^{+}-\vec{\Phi}^{-})\fracslash(2\epsilon)$\;

        $\mat{G}[i, j] \gets$ \interp{$\rvar_i$, $\vec{v}_i$, $\vec{\delta}$}\;
      }
    }

    \BlankLine

    \tcp{
      solve linear system $
        \nabla_{\parvec}\rvec
        = - (\nabla_{\rvec}\pseudocdf)\inv \cdot \nabla_{\parvec}\pseudocdf
    $}

    $\mat{J} \gets$ \linsolve{$\mat{H}$, $\mat{G}$}\;
      \nllabel{alg:get_sensitivity_nd:inv}

    \KwRet $\mat{J}$
  }
\end{algorithm}

\Cref{alg:get_sensitivity_nd_grid} instead solves the linear systems at all grid vertices in $\vec{V}$ and interpolates to queried points.
Thus the number of calls to $\density$ is independent of the number of query points.
It minimizes those calls by reusing joint PDF values at vertices via \getconditionalsText (\Cref{alg:get_conditionals}).
Finite-difference schemes are stated in \cref{eqn:forward_difference,eqn:backward_difference,eqn:central_difference}.

\begin{algorithm}[htbp!]
  \SetKwFunction{gradndgrid}{get\_sensitivity\_nd\_grid}
  \caption{%
    The second numerical algorithm for evaluating \cref{eqn:nd_sensitivity_def}. %
    Later on in \Cref{sec:benchmarks}, it is labeled as \InterpFull. %
    The external function \getconditionalsText is described in \Cref{alg:get_conditionals} in \ref{sec:algorithms}. %
    Finite difference schemes for \texttt{forward}, \texttt{backward}, and \texttt{central} are described in \cref{eqn:forward_difference,eqn:backward_difference,eqn:central_difference}, respectively.%
  }
  \label{alg:get_sensitivity_nd_grid}

  \KwIn{$\density$, $\vec{V}$, $\parvec$, $\mat{X}$, $\epsilon$}

  \KwOut{%
    $\mat{J}$ ($\mat{J} \in \Rnum{M\times\nrvars\times\npars}$; %
    Jacobian matrices for all $M$ points in $\mat{X}$)%
  }

  \BlankLine

  \Fn{\gradndgrid{$\density$, $\vec{V}$, $\parvec$, $\mat{X}$, $\epsilon$}}{

    \BlankLine

    \tcp{initialize arrays}
    $\mat{J}_{v} \gets
      \mathbf{0}_{K_1 \times \dots \times K_\nrvars \times \nrvars \times \npars}$
      \tcp*[l]{for vertices}
    $\mat{J} \gets
      \mathbf{0}_{M \times \nrvars \times \npars}$
      \tcp*[l]{for points in $\mat{X}$}

    \BlankLine

    \tcp{conditional CDFs at vertices under the given parameters}
    $(\mat{\Phi}_1, \ldots, \mat{\Phi}_\nrvars) \gets$
      \getconditionals{$\density$, $\vec{V}$, $\parvec$}[2]\;

    \BlankLine

    \tcp{conditional CDFs at vertices under perturbed parameters}
    \For(){$j \gets 1$ \KwTo $\npars$}{

      $(\mat{\Phi}_{1j}^{+}, \mat{\Phi}_{2j}^{+}, \ldots, \mat{\Phi}_{\nrvars j}^{+})
        \gets$
        \getconditionals{$\density$, $\vec{V}$, $\parvec+\epsilon\vec{e}_j$}[2]\;

      $(\mat{\Phi}_{1j}^{-}, \mat{\Phi}_{2j}^{-}, \ldots, \mat{\Phi}_{\nrvars j}^{-})
        \gets$
        \getconditionals{$\density$, $\vec{V}$, $\parvec-\epsilon\vec{e}_j$}[2]\;
    }

    \BlankLine

    \tcp{get the sensitivity at each vertex; $\vec{k}=(k_1, \ldots, k_\nrvars)$}
    \ForEach(){$\vec{k} \in \prod_{l=1}^{\nrvars} \{1, \ldots, K_l\}$}{

      \BlankLine

      $\mat{H} \gets
        \mathbf{0}_{\nrvars \times \nrvars}$
        \tcp*[l]{$\mat{H}\coloneqq\nabla_{\rvec}\pseudocdf$ at vertex $\vec{k}$}

      $\mat{G} \gets
        \mathbf{0}_{\nrvars \times \npars}$
        \tcp*[l]{$\mat{G}\coloneqq\nabla_{\parvec}\pseudocdf$ at vertex $\vec{k}$}

      \BlankLine

      \For(\tcp*[h]{loop over each conditional}){$i \gets 1$ \KwTo $\nrvars$}{

        \BlankLine

        \For(\tcp*[h]{loop over each spatial coordinate}){$j \gets 1$ \KwTo $\nrvars$}{

          \If(\tcp*[h]{1st vertex; \cref{eqn:forward_difference}}){$k_j = 1$}{
            $\vec{k}^{+1},\,\vec{k}^{+2} \gets
              (k_1, \ldots, k_j+1, \ldots, k_\nrvars),\,
              (k_1, \ldots, k_j+2, \ldots, k_\nrvars)$\;
            $\mat{H}[i, j] \gets \mathtt{forward}(
              v_j^{k_{j}}, v_j^{k_{j}+1}, v_j^{k_{j}+2},
              \mat{\Phi}_i[\vec{k}],
              \mat{\Phi}_i[\vec{k}^{+1}],
              \mat{\Phi}_i[\vec{k}^{+2}]
            )$\;
          }
          \ElseIf(\tcp*[h]{last vertex; \cref{eqn:backward_difference}}){$k_j = K_j$}{
            $\vec{k}^{-2},\,\vec{k}^{-1} \gets
              (k_1, \ldots, k_j-2, \ldots, k_\nrvars),\,
              (k_1, \ldots, k_j-1, \ldots, k_\nrvars)$\;
            $\mat{H}[i, j] \gets \mathtt{backward}(
              v_j^{k_{j}-2}, v_j^{k_{j}-1}, v_j^{k_{j}},
              \mat{\Phi}_i[\vec{k}^{-2}],
              \mat{\Phi}_i[\vec{k}^{-1}],
              \mat{\Phi}_i[\vec{k}]
            )$\;
          }
          \Else(\tcp*[h]{rest vertices; \cref{eqn:central_difference}}){
            $\vec{k}^{-},\,\vec{k}^{+} \gets
              (k_1, \ldots, k_j-1, \ldots, k_\nrvars),\,
              (k_1, \ldots, k_j+1, \ldots, k_\nrvars)$\;
            $\mat{H}[i, j] \gets \mathtt{central}(
              v_j^{k_{j}-1}, v_j^{k_{j}}, v_j^{k_{j}+1},
              \mat{\Phi}_i[\vec{k}^{-}],
              \mat{\Phi}_i[\vec{k}],
              \mat{\Phi}_i[\vec{k}^{+}])
            $\;
          }
        }

        \BlankLine

        \For(\tcp*[h]{loop over each parameter}){$j \gets 1$ \KwTo $\npars$}{
          $\mat{G}[i, j] \gets
            (\mat{\Phi}_{ij}^{+}[\vec{k}] - \mat{\Phi}_{ij}^{-}[\vec{k}])
            \fracslash (2\epsilon)$\;
        }
      }

      \BlankLine

      \tcp{%
        solve linear system $\nabla_{\parvec}\rvec = - (\nabla_{\rvec}\pseudocdf)\inv
        \cdot \nabla_{\parvec}\pseudocdf$ for vertex $\vec{k}$%
      }
      $\mat{J}_{v}[\vec{k}, :, :] \gets$
        $(-1) \times$\linsolve{$\mat{H}$,
        $\mat{G}$}\tcp*[l]{$\mat{J_v}[\vec{k}, :, :] \in \Rnum{\nrvars\times\npars}$}
        \;
    }

    \BlankLine
    \tcp{interpolate the Jacobian at the desired space points $\mat{X}[i, :]$}
    \For(){$i \gets 1$ \KwTo $M$}{
      $\mat{J}[i, :, :] \gets$ \interpnd{$\mat{X}[i, :]$, $\mat{V}$, $\mat{J}_{v}$}
      \tcp*[l]{N-D interpolation}
    }

    \KwRet $\mat{J}$\;
  }
\end{algorithm}

\subsubsection{On the Global Bijectivity Assumption.}
\label{subsubsec:nd_bijectivity}

In practice, breakdowns in invertibility of \eqref{eqn:nd_cdf} are easy to detect by monitoring $\nabla_{\rvec} \pseudocdf$ (e.g., determinant, conditioning, or rank tests).
If detected, switch to the provably bijective, but potentially more expensive, Rosenblatt mapping.

The Rosenblatt mapping constructs a triangular sequence of marginal and conditional CDFs.
\begin{subequations}
  \begin{equation}
    \pseudocdf(\rvec;\,\parvec)
    =
    \left[
        F_1(\rvar_1;\,\parvec),\,
        F_2(\rvar_2;\,\parvec),\,
        F_3(\rvar_3;\,\parvec),\,
        \dots,\,
        F_N(\rvar_N;\,\parvec)
    \right]\tran
    =
    \vec{u}
    \,,
  \label{eqn:Rosenblatt_Transform_1}
  \end{equation}
  \text{where}
  \begin{equation}
    F_i(x_i;\,\parvec)
    \coloneq
    F_i(\rvar_i \condon \rvec_{1:i-1};\,\parvec)
    =
    \int\limits_{z=-\infty}^{z=\rvar_i}
      \density_i(z \condon \rvec_{1:i-1};\,\parvec)
      \diff z
    =
    u_i
  \label{eqn:Rosenblatt_Transform_2}
  \end{equation}
  \text{and}
  \begin{equation}
    \density_i(x_i \condon \rvec_{1:i-1};\,\parvec)
    =
    \frac{
      \int\limits_{x_{i+1}} \cdots \int\limits_{x_n}
      f(\rvec;\,\parvec)\,
      \diff x_{i+1} \cdots \diff x_{n}
    }{
      \int\limits_{x_{i}} \int\limits_{x_{i+1}} \cdots \int\limits_{x_n}
      f(\rvec;\,\parvec)\,
      \diff x_{i} \diff x_{i+1} \cdots \diff x_{n}
    }
  \label{eqn:Rosenblatt_Transform_3}
  \end{equation}
  \label{eqn:Rosenblatt_Transform}
\end{subequations}
Here, $\rvec_{1:i-1}$ denotes $\left[\rvar_1,\, \ldots,\, \rvar_{i-1}\right]$, with the convention that $\rvec_{1:0}$ is an empty vector in $\density_1$ and $\cdf_1$.

The mapping \eqref{eqn:Rosenblatt_Transform} is guaranteed to be a bijection between $\support\rvec$ and $[0,1]^N$; see, for example, \cite{rosenblatt_remarks_1952}.
Since it requires evaluating nested conditionals, however, this mapping may incur a higher cost than the simpler 1-D conditional CDFs in \eqref{eqn:nd_cdf} under a direct adaptation of the present algorithms.
Nevertheless, \eqref{eqn:Rosenblatt_Transform} can serve as a drop-in replacement in \eqref{eqn:nd_sensitivity_def} whenever bijectivity must be enforced.

\subsection{Diagonal Approximation}
\label{subsec:inverse_mat_approx}

For applications that do not require highly accurate sensitivities, we employ a diagonal approximation to $\nabla_{\rvec} \pseudocdf$ in \eqref{eqn:nd_sensitivity_def}:
\begin{equation}
  \nabla_{\rvec} \pseudocdf
  \approx
  \mathbf{Diag}\left(
    \left[
        \pderiv{\cdf_1}{\rvar_1},\, \ldots,\, \pderiv{\cdf_\nrvars}{\rvar_\nrvars}
    \right]
  \right)
  =
  \mathbf{Diag}\left(
    \left[
        \density_1,\, \ldots,\, \density_\nrvars
    \right]
  \right)
  \label{eqn:inv_mat_approx}
  \,,
\end{equation}
where $\mathbf{Diag}\left(\vec{a}\right)$ denotes a diagonal matrix with the vector $\vec{a}$ on its main diagonal.
Substituting \eqref{eqn:inv_mat_approx} into \eqref{eqn:nd_sensitivity_def}, we obtain the following approximation for the space-parameter sensitivity $\nabla_{\parvec}\rvec$:
\begin{equation}
  \nabla_{\parvec} \rvec
  \approx
  \left[
    -\frac{1}{\density_1} \pderiv{\cdf_1}{\parvec} ,\,
    \ldots ,\,
    -\frac{1}{\density_\nrvars} \pderiv{\cdf_{\nrvars}}{\parvec}
  \right]\tran
  \label{eqn:nd_sensitivity_approx}
  \,.
\end{equation}

This approximation can be inaccurate when $\nabla_{\rvec} \pseudocdf$ is not diagonally dominant, affecting the accuracy of $\nabla_{\parvec} \rvec$.
Nevertheless, for many optimization tasks---where the gradients affect search directions but are not the directions themselves, fast approximate gradients are acceptable and can reduce time-to-solution by lowering per-iteration cost.

Similar to \Cref{subsec:method_nd}, we present two solution algorithms for evaluating \eqref{eqn:nd_sensitivity_approx}: \Cref{alg:get_sensitivity_nd_diag} (labeled as \DiagApprox in \Cref{sec:benchmarks}) and \Cref{alg:get_sensitivity_nd_grid_diag} (labeled as \InterpDiag).
\Cref{alg:get_sensitivity_nd_diag} evaluates the sensitivity at each desired spatial point.
The computational cost scales linearly with the number of desired spatial points.
In \Cref{alg:get_sensitivity_nd_diag} we reuse the \getsensitivityonedText function defined in \Cref{alg:get_sensitivity_1d}.
This reuse is possible because each row in \eqref{eqn:nd_sensitivity_approx} corresponds directly to the 1-D formula in \eqref{eqn:1d_sensitivity_def}, with the 1-D distribution replaced by the $i$th conditional.
This structural similarity makes the implementation of the diagonal approximation relatively simpler than the algorithms in \Cref{subsec:method_nd}.
On the other hand, \Cref{alg:get_sensitivity_nd_grid_diag} evaluates the sensitivity at vertices of an N-D grid and interpolates the results to all desired spatial points.
The computational cost for \Cref{alg:get_sensitivity_nd_grid_diag} is thus independent of the size of queried points.

\begin{algorithm}[htbp!]
  \SetKwFunction{gradnddiag}{get\_sensitivity\_nd\_diag}
  \caption{%
    The first algorithm for evaluating \cref{eqn:nd_sensitivity_approx}. %
    It is later labeled as \DiagApprox in \Cref{sec:benchmarks}. %
    The external function \getsensitivityonedText is described in \Cref{alg:get_sensitivity_1d}.%
  }
  \label{alg:get_sensitivity_nd_diag}

  \KwIn{$\density$, $\vec{V}$, $\parvec$, $\rvec$, $\epsilon$}

  \KwOut{%
    $\mat{J}$ ($\mat{J} \coloneqq \nabla_{\parvec} \rvec$ %
    as in \cref{eqn:nd_sensitivity_approx})%
  }

  \BlankLine
  \Fn{\gradnddiag{$\density$, $\vec{V}$, $\parvec$, $\rvec$, $\epsilon$}}{

    $\mat{J} \gets \mathbf{0}_{\nrvars \times \npars}$\;

    \For(\tcp*[h]{$i$-th conditional}){$i \gets 1$ \KwTo $\nrvars$}{

      $\vec{e}_i \gets$ $i$-th standard basis in $\Rnum{\nrvars}$\;

      $\density_i(s,\,\parvec) \gets
        \density(\rvec + (s - \rvar_i) \vec{e}_i,\,\parvec)$
        \tcp*[l]{a lambda/local function}

      $\mat{J}[i, :] \gets$
        \gradonedim{$\density_i$, $\vec{v}_i$, $\parvec$, $\rvar_i$, $\epsilon$}
        \;
    }

    \KwRet $\mat{J}$\;
  }
\end{algorithm}

\begin{algorithm}[htbp!]
  \SetKwFunction{gradndgriddiag}{get\_sensitivity\_nd\_grid\_diag}
  \caption{%
    The second algorithm for evaluating \cref{eqn:nd_sensitivity_approx}. %
    It is later labeled as \InterpDiag in \Cref{sec:benchmarks}. %
    The external function \getconditionalsText is described in \Cref{alg:get_conditionals}.%
  }
  \label{alg:get_sensitivity_nd_grid_diag}

  \KwIn{$\density$, $\vec{V}$, $\parvec$, $\mat{X}$, $\epsilon$}

  \KwOut{%
    $\mat{J}$ ($\mat{J} \in \Rnum{M\times\nrvars\times\npars}$; %
    Jacobian matrices for all $M$ points in $\mat{X}$)%
  }

  \BlankLine
  \Fn{\gradndgriddiag{$\density$, $\vec{V}$, $\parvec$, $\mat{X}$, $\epsilon$}}{

    \BlankLine
    \tcp{conditional PDFs at vertices under the given parameters}
    $(\mat{\phi}_1, \ldots, \mat{\phi}_\nrvars) \gets$
      \getconditionals{$\density$, $\vec{V}$, $\parvec$}[1]
      \;

    \BlankLine
    \tcp{conditional CDFs at vertices under perturbed parameters}
    \For(){$j \gets 1$ \KwTo $\npars$}{
      $(\mat{\Phi}_{1j}^{+}, \mat{\Phi}_{2j}^{+}, \ldots, \mat{\Phi}_{\nrvars j}^{+})
        \gets$
        \getconditionals{$\density$, $\vec{V}$, $\parvec+\epsilon\vec{e}_j$}[2]
        \;

      $(\mat{\Phi}_{1j}^{-}, \mat{\Phi}_{2j}^{-}, \ldots, \mat{\Phi}_{\nrvars j}^{-})
        \gets$
        \getconditionals{$\density$, $\vec{V}$, $\parvec-\epsilon\vec{e}_j$}[2]
        \;
    }

    \BlankLine
    \For(\tcp*[h]{loop over conditionals}){$i \gets 1$ \KwTo $\nrvars$}{
      \For(\tcp*[h]{loop over each parameter}){$j \gets 1$ \KwTo $\npars$}{

        \BlankLine
        \tcp{%
          $\mat{\Delta} \in \Rnum{K_1 \times \dots \times K_\nrvars}$ %
          is $\pderivi{\cdf_i}{\param_j}$ at all vertices%
        }
        $\mat{\Delta} \gets
          (\mat{\Phi}_{ij}^{+} - \mat{\Phi}_{ij}^{-}) \fracslash (2\epsilon)$
          \tcp*[l]{element-wise subtraction \& scaling}

        \BlankLine
        \tcp{interpolation}
        \For(){$m \gets 1$ \KwTo $M$}{

          $\density_x \gets \interpnd{$\mat{X}[m, :]$, $\mat{V}$, $\mat{\phi}_{i}$}$
            \tcp*[l]{conditional PDF}

          $\delta \gets \interpnd{$\mat{X}[m, :]$, $\mat{V}$, $\mat{\Delta}$}$
            \tcp*[l]{$\pderivi{\cdf_i}{\param_j}$}

          $\mat{J}[m, i, j] \gets -\, \delta \fracslash \density_x$
            \tcp*[l]{%
              $\pderivi{\rvar_i}{\param_j}= - (\pderivi{\cdf_i}{\param_j})
              \fracslash
              \density_i$ as in \cref{eqn:nd_sensitivity_approx}
            }
        }
      }
    }

    \BlankLine
    \KwRet $\mat{J}$\;
  }
\end{algorithm}

\subsection{On Multiple Disjoint Intervals}
\label{subsec:when_f_ge_zero}

Although we assumed $\density > 0$, the method extends to continuous $\density \ge 0$ by showing the mapping $(\vec{u}, \parvec) \mapsto \rvec$ exists for all realizations.

Consider a general 1-D continuous distribution with PDF $\density(\rvar;\,\parvec) \ge 0$ defined on the domain $\domain{\density}$.
We decompose $\domain{\density}$ into the support $\support{\rvar}$ and the zero-density region $\domain{0}$, that is, $\domain{0} = \{\rvar \in \domain{\density} \mid \density(\rvar;\,\parvec) = 0\}$ and $\support{\rvar} = \{\rvar \in \domain{\density} \mid \density(\rvar;\,\parvec) > 0\}$.
By definition, $\support{\rvar}$ and $\domain{0}$ are disjoint but adjacent because $\density$ is continuous.
For simplicity, assume $\domain{0}$ is a single interval; the same reasoning applies to multiple disjoint intervals.
In 1-D, the CDF is non-decreasing and can be expressed in a piecewise manner as
\begin{equation}
  \cdf(\rvar;\,\parvec) =
  \begin{cases}
    \cdf_{0} = c & \text{if } \rvar \in \domain{0}\\
    \cdf_{\ne 0} = u \in [0, 1] \setminus \{c\} & \text{if } \rvar \in \support{\rvar}\\
  \end{cases}
  \label{eqn:piecewise_cdf}
  \,.
\end{equation}
Here, $c \in [0, 1]$ is a constant; since no realizations exist in $\domain{0}$, the cumulative probability remains constant on this interval.
For $\rvar \in \support{\rvar}$, the CDF takes values between 0 and 1, excluding $c$.

Note that exactly one point $\rvar$ in the closure of $\support{\rvar}$ (i.e., $\overline{\support{\rvar}}$) satisfies $\cdf(\rvar;\,\parvec) = c$.
Because $\density$ is continuous, the boundary point separating $\domain{0}$ and $\support{\rvar}$ belongs to $\domain{0}$, except for a set of measure zero.
Most importantly, while $\cdf_{0}$ is not a one-to-one mapping, $\cdf_{\ne 0}$ is, which ensures that the inverse mapping exists for all $u \in [0, 1]$ except at $u = c$, as shown in the piecewise inverse CDF:
\begin{equation}
  \cdf\inv(u;\,\parvec) =
  \begin{cases}
    \text{undefined} & \text{if }u = c\\
    \cdf_{\ne 0}\inv(u;\,\parvec) = \rvar \in \support{\rvar} & \text{otherwise}
  \end{cases}
  \label{eqn:piecewise_invcdf}
  \,.
\end{equation}

Equations~\eqref{eqn:piecewise_cdf} and \eqref{eqn:piecewise_invcdf} state that the inverse mapping is well defined for all $\rvar \in \support{\rvar}$.
Recall that our work only considers sensitivity estimation at realizations $\rvar \in \support{\rvar}$.
This implies that we do not encounter values of $u$ equal to $c$.
Thus, the existence of $\cdf_{\ne 0}\inv$ is sufficient for our method to be applicable.
Accordingly, for the general case where $\density \ge 0$, we adopt the notation $\cdf$ and $\cdf\inv$ to implicitly refer to $\cdf_{\ne 0}$ and $\cdf_{\ne 0}\inv$, respectively.
For N-D cases, the same reasoning applies, since our method operates on 1-D conditional CDFs.

\subsection{Computational Cost and Other Considerations}
\label{subsec:comp_cost}

We now estimate costs, focusing on calls to the joint PDF $\density$, which is often the dominant computational expense in sample-based inference (e.g., \Cref{subsec:app_dis}).

Consider a multivariate distribution with $\nrvars$ spatial dimensions and $\npars$ distribution parameters, a rectilinear grid with $K_i$ vertices in the $i$th spatial dimension, and $M$ spatial points (i.e., realizations) at which we want to evaluate the sensitivities.
\Cref{tab:comp_cost} lists the total number of function calls to $\density$ for each algorithm.

\begin{table}[H]
  \centering
  \caption{Number of function calls to $\density$ for each algorithm.}
  \label{tab:comp_cost}
  \small
  \renewcommand{\arraystretch}{1.2}
  \begin{tabularx}{\textwidth}{cccX}
    \toprule
    Equation & Algorithm ID & Label in \Cref{sec:benchmarks} & Number of Function Calls
    \\
    \midrule
    \eqref{eqn:nd_sensitivity_def} & \Cref{alg:get_sensitivity_nd}
      & \FullInv
      & $2M(\nrvars+\npars)\sum_{i=1}^{\nrvars} K_i$
    \\
    \eqref{eqn:nd_sensitivity_def} & \Cref{alg:get_sensitivity_nd_grid}
      & \InterpFull
      & $(2\npars+1)\prod_{i=1}^{\nrvars} K_i$
    \\
    \eqref{eqn:nd_sensitivity_approx} & \Cref{alg:get_sensitivity_nd_diag}
      & \DiagApprox
      & $M\left[\nrvars + \left(2\npars+1\right)\right]\sum_{i=1}^{\nrvars} K_i$
    \\
    \eqref{eqn:nd_sensitivity_approx} & \Cref{alg:get_sensitivity_nd_grid_diag}
      & \InterpDiag
      & $(2\npars+1)\prod_{i=1}^{\nrvars} K_i$
    \\
    \bottomrule
   \end{tabularx}
 \end{table}

For \cref{eqn:nd_sensitivity_def}, \Cref{alg:get_sensitivity_nd_grid} is more efficient than \Cref{alg:get_sensitivity_nd} when $M > \frac{(2\npars+1)\prod_{i=1}^{\nrvars} K_i} {2(\nrvars+\npars)\sum_{i=1}^{\nrvars} K_i}$.
For example, with $\nrvars=3$, $\npars=256$, and $K_1=K_2=K_3=128$, the threshold is $\num{5408}$ points.
Similarly, for \eqref{eqn:nd_sensitivity_approx}, \Cref{alg:get_sensitivity_nd_grid_diag} is more efficient than \Cref{alg:get_sensitivity_nd_diag} when $M > \frac{(2\npars+1)\prod_{i=1}^{\nrvars} K_i} {\nrvars + (2\npars+1) \sum_{i=1}^{\nrvars} K_i}$ (threshold $\num{5461}$ in the same example).

We do not compare the costs of evaluating \eqref{eqn:nd_sensitivity_def} versus evaluating \eqref{eqn:nd_sensitivity_approx}, as they are different mathematical approximations; the choice of which equations to evaluate should be driven by application needs.

Beyond function calls, \Cref{alg:get_sensitivity_nd_grid} and \Cref{alg:get_sensitivity_nd_grid_diag} are more memory-demanding than their pointwise counterparts due to storing values at all vertices for piecewise multilinear interpolation~\cite{weiser_note_1988}.
When peak memory is a concern, \Cref{alg:get_sensitivity_nd} and \Cref{alg:get_sensitivity_nd_diag} may be preferable.
Advanced numerical schemes (higher-order basis, tailored meshes, or mesh-free methods) and code refactoring (e.g., loop reordering to reduce temporary storage of $\Phi_{ij}^{\pm}$ in \Cref{alg:get_sensitivity_nd_grid_diag}) can alleviate memory pressure.


\section{Verification, Validation, and Benchmarks}
\label{sec:benchmarks}

We provide three groups of numerical experiments for the purposes of verification and validation (V\&V~\cite{oberkampf_verification_2010}):
\begin{itemize}
  \item \textbf{Verification} (\Cref{subsec:verification}):
    We verify that the numerical algorithms described in \Cref{sec:methods} properly approximate the proposed formulae \eqref{eqn:1d_sensitivity_def}, \eqref{eqn:nd_sensitivity_def}, and \eqref{eqn:nd_sensitivity_approx}.
    Specifically, we compare our numerical results with closed forms of the three formulae for 1-D and 2-D Gaussian distributions.
    These closed forms are derived in \ref{sec:analytical_gaussian}.
  \item \textbf{Validation} (\Cref{subsec:validation}):
    We validate that the proposed formulae \eqref{eqn:1d_sensitivity_def}, \eqref{eqn:nd_sensitivity_def}, and \eqref{eqn:nd_sensitivity_approx} yield the sensitivity $\nabla_{\parvec} \rvec$ that is effective for sample-based inference.
    Specifically, we consider the energy score $L$, a widely used loss function for sample-based inference~\cite{gneiting_strictly_2007, constantinescu_statistical_2020}.
    Numerical results of $\nabla_{\parvec}L$, which require $\nabla_{\parvec} \rvec$ through the relation $\nabla_{\parvec} L = \sum_{k} \left(\nabla_{\parvec} \rvec_k\right)^\mathsf{T} \nabla_{\rvec_k} L$, are compared with the analytical results of $\nabla_{\parvec}L$.
    We also perform sample-based inference to assess both correctness and computational efficiency.
    The selected tests are a 1-D beta distribution and a 2-D distribution commonly used in nuclear physics benchmarking.
  \item \textbf{Benchmarking} (\Cref{subsec:app_dis}):
    We demonstrate that our proposed method effectively resolves a realistic inference problem in nuclear physics, where the PDF is non-analytical and requires complex forward physics simulations.
\end{itemize}

For our convenience, we briefly summarize in \Cref{tab:alg_list} all the algorithms.
The table columns provide the corresponding equation that an algorithm evaluates (first column), the locations of algorithmic statements (second column), meaningful algorithm names for improved readability (third column), and notes on applicability and differences from similar algorithms (fourth column).

\begin{table}[htbp!]
  \centering
  \caption{
    Algorithm labels, pseudocode locations, and approximated formulae.
  }
  \label{tab:alg_list}
  \small
  \renewcommand{\arraystretch}{1.2}
  \begin{tabularx}{\textwidth}{cccX}
    \toprule
    Eq. & Pseudocode & Label & Description
    \\
    \midrule
    \eqref{eqn:1d_sensitivity_def} & \Cref{alg:get_sensitivity_1d}
      & \OneDAlg
      & Applicable only to 1-D distributions.
    \\
    \eqref{eqn:nd_sensitivity_def} & \Cref{alg:get_sensitivity_nd}
      & \FullInv
      & For arbitrary dimensionality.
        Solves $\nabla_{\parvec}\rvec$ at all given realizations.
    \\
    \eqref{eqn:nd_sensitivity_def} & \Cref{alg:get_sensitivity_nd_grid}
      & \InterpFull
      & For arbitrary dimensionality.
        Solves $\nabla_{\parvec}\rvec$ only at Cartesian grid vertices and interpolates to realizations.
    \\
    \eqref{eqn:nd_sensitivity_approx} & \Cref{alg:get_sensitivity_nd_diag}
      & \DiagApprox
      & For arbitrary dimensionality.
        Solves $\nabla_{\parvec}\rvec$ at all given realizations.
    \\
    \eqref{eqn:nd_sensitivity_approx} & \Cref{alg:get_sensitivity_nd_grid_diag}
      & \InterpDiag
      & For arbitrary dimensionality.
        Solves $\nabla_{\parvec}\rvec$ only at Cartesian grid vertices and interpolates to realizations.
    \\
    \bottomrule
   \end{tabularx}
 \end{table}

\subsection{Verification: Analytical Sensitivities of 1-D and 2-D Gaussian Distributions}
\label{subsec:verification}

\subsubsection{1-D Gaussian Distribution.}
\label{subsubsec:verification_Gaussian_1d}

The sensitivity of a 1-D Gaussian random variable $\rvar \sim \mathcal{N}(\mu, \sigma^2)$ with respect to its parameters $\parvec \coloneqq \left[\mu,\, \sigma\right]\tran$ is derived in \ref{subsec:analytical_1d_gaussian} and is given by
\begin{equation}
  \nabla_{\parvec} \rvar
  =
  \left[ 1 ,\,  \frac{(x - \mu)}{\sigma} \right]\tran
  \,.
  \label{eqn:1d_gauss_grad}
\end{equation}

While a Gaussian distribution has infinite support, all our numerical algorithms require a finite domain to numerically compute the CDF $\cdf$ and its derivatives.
We therefore define the computational domain as $\mu \pm 5\sigma$ and the domain of interest as $\mu \pm 4\sigma$.
This approach approximates the 1-D Gaussian distribution with the finite computational domain, which is equivalent to treating probability outside $\mu\pm 5\sigma$ as zero.
The resulting approximation error is approximately $10^{-6}$, corresponding to the probability mass of a 1-D Gaussian distribution lying outside $\mu \pm 5\sigma$.
The narrower domain of interest ensures accurate sensitivity computations within this region, while the boundary zones $[\mu-5\sigma, \mu-4\sigma]$ and $[\mu+4\sigma, \mu+5\sigma]$ serve as buffers that mitigate finite-domain boundary effects.

\begin{figure}[htbp!]
  \centering
    \includegraphics[width=0.495\linewidth]{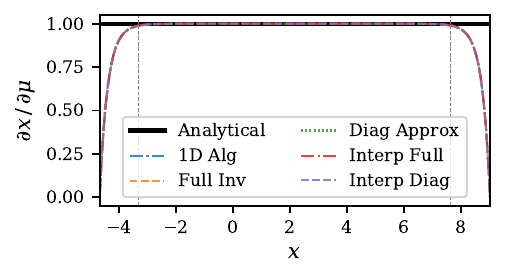}
    \includegraphics[width=0.495\linewidth]{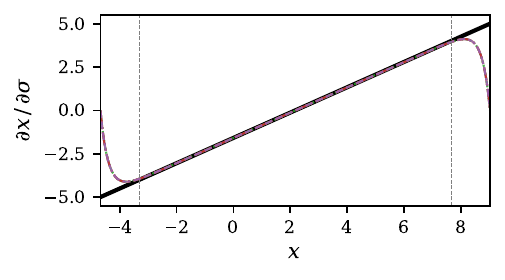}
  \caption{%
    Comparison of analytical and numerical sensitivities for a 1-D Gaussian %
    distribution with parameters $\mu = 2.175$ and $\sigma = 1.371$. %
    Left plot shows $\nabla_{\mu} \rvar$; right plot shows $\nabla_{\sigma} \rvar$. %
    The computational domain spans $\mu \pm 5\sigma$, with the domain of interest %
    $\mu \pm 4\sigma$ marked by vertical dashed lines. %
    Boundary effects can be seen in the boundary intervals $[\mu-5\sigma, \mu-4\sigma]$ and $[\mu+4\sigma, \mu+5\sigma]$. %
  }
  \label{fig:1d_gaussian_sensitivity}
\end{figure}

\Cref{fig:1d_gaussian_sensitivity} shows the sensitivity results using $\mu = 2.175$ and $\sigma = 1.371$.
All the algorithms produce visually identical results, as expected for this 1-D case.
Since both \eqref{eqn:nd_sensitivity_def} (solved by \FullInv and \InterpFull) and \eqref{eqn:nd_sensitivity_approx} (solved by \DiagApprox and \InterpDiag) reduce to \eqref{eqn:1d_sensitivity_def} for 1-D distributions, all the algorithms should yield near-identical results.
Negligible differences may arise from variations in computational procedures and the additional interpolations used by \InterpFull and \InterpDiag.

As shown in \Cref{fig:1d_gaussian_sensitivity}, deviations from the analytical values occur near the domain boundaries.
This effect arises from approximating the infinite-domain Gaussian distribution on a finite computational domain.
In a 1-D finite domain distribution, the normalized CDF $F$ at the domain boundaries is always $0$ and $1$, respectively, because the CDF calculation accumulates the probability mass from the lower bound to the upper bound of the domain.
Consequently, the sensitivity at the boundaries is always zero in 1-D, since $\frac{1}{\density} \pderiv{\cdf}{\parvec} = \frac{1}{\density} \cdot 0 = 0$.
The closer a point is to the boundary, the more pronounced this domain effect becomes.
This behavior does not exist in the analytical sensitivity \eqref{eqn:1d_gauss_grad} for the original Gaussian distribution, since it is defined over an infinite domain.
Therefore, in practice, one should use a computational domain larger than the domain of interest to mitigate boundary effects when working with unbounded distributions.

All our numerical algorithms use grid-based schemes (for differentiation, integration, and interpolation) for approximating $\cdf$ and its derivatives.
We thus distinguish between two types of grids in this work.
The \emph{background grid} refers to the grid used for numerically approximating $\cdf$ and its derivatives and is independent of where we evaluate the sensitivity.
In contrast, the \emph{foreground grid} refers to the grid of evaluation points where we compute the sensitivity whenever we need to place evaluation points in a grid-like pattern.
For example, when evaluating numerical error via \cref{eqn:l1_error}, we require a foreground grid for the trapezoidal rule.

To examine how the resolution of the background grid affects sensitivity accuracy, we run each algorithm using uniform background grids with varying numbers of vertices $N$ over the computational domain $\mu \pm 5\sigma$.
We use the $L_1$ error as the metric,
\begin{equation}
  L_1
  =
  \lim\limits_{M \to \infty}
  \frac{1}{M} \sum_{k=1}^{M} \lvert c(x_k) - a(x_k)\rvert
  =
  \int_{\mu - 4\sigma}^{\mu + 4\sigma}
  \lvert c(x) - a(x) \rvert
  \density(\rvar) \diff\rvar
  \,.
  \label{eqn:l1_error}
\end{equation}
Here, $c(x)$ and $a(x)$ denote the numerical and analytical sensitivities at location $x$, respectively.
The $L_1$ error is defined as the expected absolute error between numerical and analytical sensitivities.
We used the trapezoidal rule to evaluate the integral in \eqref{eqn:l1_error} using $2^{14}$ uniformly spaced points for the foreground grid within the domain of interest.

For all numerical derivatives of $\cdf$ in all the algorithms, we use finite differences with a fixed step size of $\epsilon = 10^{-5}$.

\begin{figure}[htbp!]
  \centering
    \includegraphics[width=0.495\linewidth]{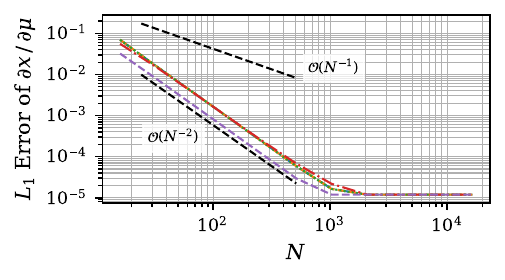}
    \includegraphics[width=0.495\linewidth]{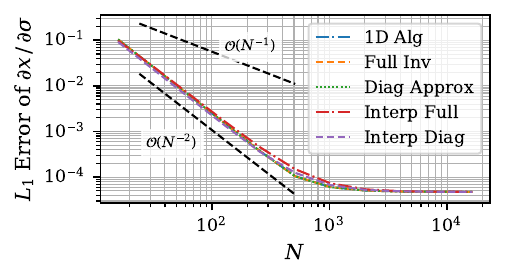}
  \caption{%
    $L_1$ errors of the computed $\nabla_{\parvec} \rvar$ compared with analytical values for a 1-D Gaussian with $\mu = 2.175$ and $\sigma = 1.371$. %
    $N$ denotes the total number of vertices in the uniform 1-D background grid that discretizes $\mu \pm 5\sigma$. %
    Sensitivities were evaluated at a foreground grid with $2^{14}$ uniformly spaced points in $\mu \pm 4\sigma$. %
    The $L_1$ errors were computed by trapezoidal integration over the foreground grid.%
  }
  \label{fig:1d_gaussian_error}
\end{figure}

The computed $L_1$ errors versus grid resolution $N$ are shown in \Cref{fig:1d_gaussian_error}.
The observed convergence rate is $\mathcal{O}(N^{-2})$.
Although we did not derive the theoretical convergence rate in this study, the empirical rate aligns with expectations because all adopted numerical schemes have second-order accuracy: the trapezoidal rule, second-order finite differences, and piecewise linear interpolation.
The errors are bounded at approximately $10^{-5}$.
As previously stated, the finite computational domain $\mu \pm 5\sigma$ inherently carries an error of approximately $10^{-6}$ compared with the original infinite-domain Gaussian.
Additional error sources contribute to the gap between $10^{-6}$ and $10^{-5}$, including the use of low-order numerical schemes and the choice of finite difference step size.
For a Gaussian distribution, rounding errors become more dominant near the domain boundaries because the probability density $\density$ approaches zero near the tails, magnifying numerical errors in $\frac{1}{\density}$, which appears in all algorithms.

\subsubsection{2-D Gaussian Distribution.}
\label{subsubsec:verification_Gaussian_2d}

Consider a 2-D Gaussian random vector $\rvec \coloneqq \left[\rvar_1, \rvar_2\right]\tran \sim \mathcal{N}(\vec{\mu}, \mat{\Sigma})$, where $\vec{\mu}=\left[\mu_1, \mu_2\right]\tran$ and $\mat{\Sigma} \coloneqq \left[ \begin{smallmatrix} \sigma_1^2 & \rho\sigma_1\sigma_2 \\ \rho\sigma_1\sigma_2 & \sigma_2^2 \end{smallmatrix} \right]$ denote the mean and the covariance matrix, respectively.
We can define the vector of distribution parameters as $\parvec \coloneqq \left[\mu_1,\, \mu_2,\, \sigma_1,\, \sigma_2,\, \rho \right]\tran$.
The closed form of the sensitivity \eqref{eqn:nd_sensitivity_def} is given by
\begin{equation}
  \nabla_{\parvec} \rvec
  =
  \begin{bmatrix}
    1 & 0 & z_1 & 0 & \frac{\sigma_1}{1-\rho^2} z_2
    \\
    0 & 1 & 0 & z_2 & \frac{\sigma_2}{1-\rho^2} z_1
  \end{bmatrix}
  \text{, where }
  z_1 \coloneqq \frac{x_1 - \mu_1}{\sigma_1}
  \text{ and }
  z_2 \coloneqq \frac{x_2 - \mu_2}{\sigma_2}
  \label{eqn:2d_gauss_grad}
  \,,
\end{equation}
and the closed form for calculating the sensitivity via diagonal approximation \eqref{eqn:nd_sensitivity_approx} is
\begin{equation}
  \nabla_{\parvec} \rvec
  \approx
  \begin{bmatrix}
    1 & -\rho\frac{\sigma_1}{\sigma_2} &
    z_1 & -\rho\frac{\sigma_1}{\sigma_2} z_2 &
    -\frac{\sigma_1}{1-\rho^2}\left( \rho z_1 - z_2 \right)
    \\
    -\rho\frac{\sigma_2}{\sigma_1} & 1 &
    -\rho\frac{\sigma_2}{\sigma_1} z_1 & z_2 &
    \frac{\sigma_2}{1-\rho^2}\left( z_1 - \rho z_2 \right)
  \end{bmatrix}
  \label{eqn:2d_gauss_grad_approx}
  \,.
\end{equation}
For clarity of the presentation, the derivation of \eqref{eqn:2d_gauss_grad} and \eqref{eqn:2d_gauss_grad_approx} is given in \ref{subsec:analytical_2d_gaussian}.

In this verification study we compare the numerical results of \FullInv and \InterpFull against \eqref{eqn:2d_gauss_grad} and compare \DiagApprox and \InterpDiag against \eqref{eqn:2d_gauss_grad_approx}.
The ground truth parameters are set to $\mu_1=0.7$, $\mu_2=-1.1$, $\sigma_1=2.6$, $\sigma_2=1.3$, and $\rho=0.678$.
The domain of interest is defined by a Mahalanobis distance less than or equal to $19.313$, capturing $99.9936\%$ of the realizations of the 2-D Gaussian distribution.
This domain of interest is comparable to the $\pm 4\sigma$ interval for the 1-D case in \Cref{subsubsec:verification_Gaussian_1d} in terms of probability mass.
The computational domain spans $\pm 5 $ standard deviations from the means in both directions to reduce boundary effects.

\begin{figure}[htbp!]
  \centering
  \includegraphics[width=\linewidth]{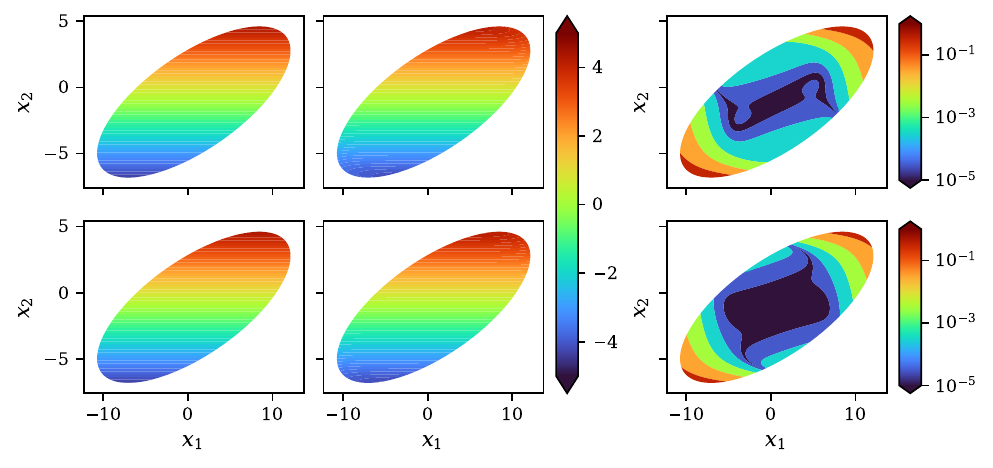}
  \caption{%
    Results for $\pderivi{\rvar_2}{\sigma_2}$.
    Top row: (Left) computed via \eqref{eqn:2d_gauss_grad}, (Center) computed via \InterpFull, (Right) absolute error between the two.
    Bottom row: (Left) computed via \eqref{eqn:2d_gauss_grad_approx}, (Center) computed via \InterpDiag, (Right) absolute error between the two.
    The frame of each plot indicates the computational domain, while only the domain of interest is visualized.
    The domain of interest is defined where the Mahalanobis distance is less than or equal to $19.313$.
    The two algorithms are compared to different analytical solutions because, without numerical error, they converge to \eqref{eqn:2d_gauss_grad} and \eqref{eqn:2d_gauss_grad_approx}, respectively.
    Thus, these plots examine the error introduced by our numerical schemes and configurations, rather than comparing the two algorithms against each other; such a comparison should instead be based on inference results, as in \Cref{fig:2d_proxy_trained_pdf_errs} in \Cref{subsec:validation}.
  }
  \label{fig:2d_gauss_res}
\end{figure}

\Cref{fig:2d_gauss_res} shows the results of $\pderivi{\rvar_2}{\sigma_2}$ using \InterpFull and \InterpDiag and their corresponding closed forms, that is, \eqref{eqn:2d_gauss_grad} and \eqref{eqn:2d_gauss_grad_approx}, respectively.
To save space and given the visual similarity of all other sensitivities $\pderivi{\rvar_i}{\alpha_i}$, we omit other figures here.
Interested readers can refer to our open-source package~\cite{attia_distrosa_2025-1} to reproduce all results and generate corresponding figures.

\Cref{fig:2d_gauss_res} shows that the computed sensitivity decreases in accuracy near the domain boundary, which is expected because of the boundary effects from approximating an infinite-domain distribution with a finite-domain distribution.
This is particularly noticeable in the upper-right and lower-left corners of the domain of interest, since these two corners are not far enough from the boundary of the computational domain.
Additionally, similar to the 1-D case, these boundary zones have near-zero probability densities, magnifying the influence of rounding errors.
Nevertheless, these regions of near-zero density correspond to rarely observed realizations and are therefore of little concern in most applications.

Note that \Cref{fig:2d_gauss_res} is not intended to compare the two algorithms directly: in the absence of numerical errors, they converge to different analytical solutions, \eqref{eqn:2d_gauss_grad} and \eqref{eqn:2d_gauss_grad_approx}, respectively, reflecting different assumptions for the gradient of a random variable.
Instead, this figure examines the errors introduced by our numerical schemes and configurations, showing that, under the same settings (e.g., domain size, interpolation order, finite difference order, grid structure, and grid spacing), \InterpDiag is closer to its corresponding analytical solution than \InterpFull is.
This difference may occur because \InterpFull performs full matrix inversions and thus introduces additional numerical factors, such as condition numbers and linear solvers, whereas \InterpDiag simply computes reciprocals of diagonal values.
A fair comparison between the two algorithms should instead be based on actual inference performance, since all algorithms are designed to make inference possible; for example, \Cref{fig:2d_proxy_trained_pdf_errs} in \Cref{subsec:validation} shows the inferred probability density functions from different algorithms.

Similar to the 1-D case, we run each algorithm with varying $N$ in the $N \times N$ background grid used for numerically approximating conditionals $\cdf_i$, their derivatives, and interpolations.
By doing so, we compute the convergence of the $L_1$ error versus $N$.
The $L_1$ error is defined similarly to \eqref{eqn:l1_error} but with the integration domain (i.e., the domain of interest) being a 2-D ellipse within the chosen Mahalanobis distance.
In order to carry out such integration, the foreground grid is configured to be a $2^{11} \times 2^{11}$ uniform grid over the computational domain (rather than over the domain of interest).
Only the evaluation points falling within the domain of interest are involved in the integration via the trapezoidal rule.

\begin{figure}[htbp!]
  \centering
  \includegraphics[width=\linewidth]{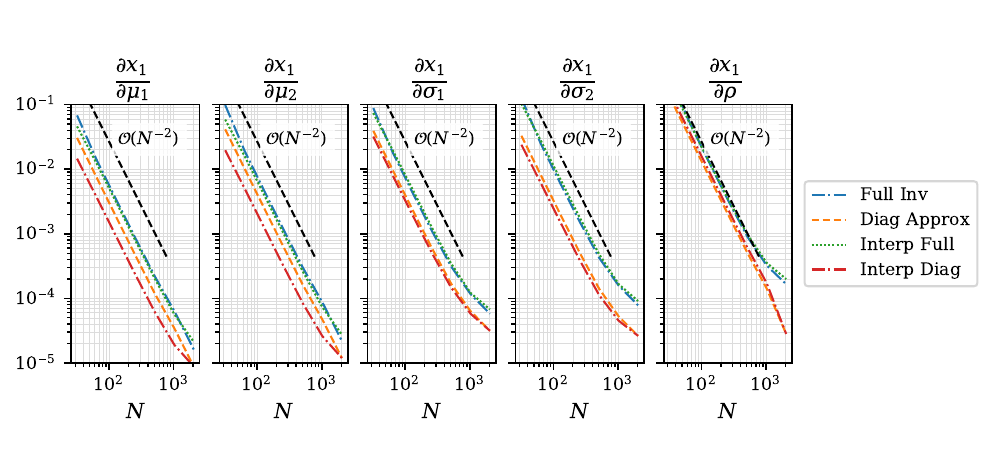}
  \caption{%
    Convergence of the $L_1$ error in $\pderivi{\rvar_1}{\parvec}$ with respect to $N$, where $N$ is defined as the square root of the total number of vertices in the background grid. %
    \FullInv and \InterpFull were compared against \eqref{eqn:2d_gauss_grad}. %
    \DiagApprox and \InterpDiag  were compared against \eqref{eqn:2d_gauss_grad_approx}.%
  }
  \label{fig:2d_gauss_converge}
\end{figure}

\Cref{fig:2d_gauss_converge} shows the convergence of $L_1$ errors with respect to $N$ for $\pderivi{\rvar_1}{\parvec}$.
We also observe a convergence rate of $\mathcal{O}(N^{-2})$ and a cap on errors around $10^{-5}$, consistent with the results of the 1-D case.
The error convergence for $\pderivi{\rvar_2}{\parvec}$ is similar, so we omit the corresponding figures here (reproducible via \cite{attia_distrosa_2025-1}).

\begin{figure}[H]
  \centering
  \includegraphics[width=0.6\linewidth]{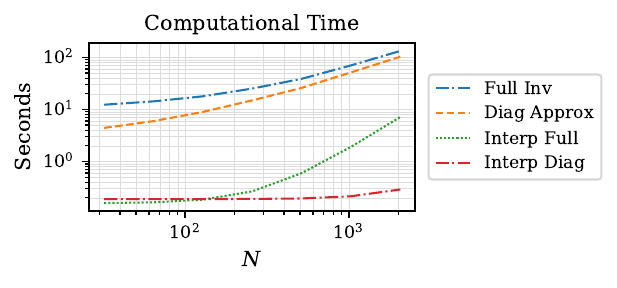}
  \caption{%
    Computational cost in wall time in seconds for the 2-D Gaussian distribution. %
    All calculations are performed on a single NVIDIA GeForce RTX 4070 GPU and double-precision floats. %
    $N$ follows the same definition as in \Cref{fig:2d_gauss_converge}.%
  }
  \label{fig:2d_gauss_comp_time}
\end{figure}

\Cref{fig:2d_gauss_comp_time} shows the computational cost in wall time in seconds.
Combining the observations in the computational cost and the error levels in \Cref{fig:2d_gauss_converge}, we note that \InterpFull and \InterpDiag cost much less but did not sacrifice accuracy significantly, indicating a better performance-cost ratio.
Note that \Cref{tab:comp_cost} counts function evaluations, whereas \Cref{fig:2d_gauss_comp_time} reports wall time that also reflects the costs of matrix inversions (in \FullInv and \InterpFull) and interpolations.
Thus, \InterpFull becomes much more expensive than \InterpDiag at larger $N$ because it inverts $N^2$ full matrices.

\subsection{Validation: Gradients of Expectation-based Loss Functions and Inference}
\label{subsec:validation}

The validation examines whether formulae \eqref{eqn:1d_sensitivity_def}, \eqref{eqn:nd_sensitivity_def}, and \eqref{eqn:nd_sensitivity_approx} yield an effective gradient of a random vector $\rvec$ in sample-based inference.
As introduced in \Cref{sec:intro}, $\nabla_{\parvec}\rvec$ is undefined without additional context.
Different contexts naturally lead to different interpretations and values for such a gradient.
For example, in our work, we interpret $\nabla_{\parvec} \rvec$ as the partial derivatives of the mapping $(\vec{u}, \parvec) \mapsto \rvec$, where $\vec{u}$ is a vector of 1-D conditional CDFs.
This interpretation differs from alternative approaches, such as methods that define $\vec{u}$ using the Rosenblatt mapping~\cite{figurnov_implicit_2018}, which would yield different values for $\nabla_{\parvec} \rvec$.
Given this context-dependence, no single ground truth exists for $\nabla_{\parvec}\rvec$ against which we can verify the correctness of our formulae.
Consequently, rather than examining whether our formulae yield a ``correct'' gradient, we focus on whether they yield an ``effective'' gradient in applications that require gradient information of $\rvec$.

We apply our formulae to two sample-based inference problems: fitting the 1-D beta distribution (\Cref{subsubsec:1d_beta_dist}) and fitting a 2-D proxy distribution commonly used in nuclear physics for benchmarking (\Cref{subsubsec:2d_proxy}).
The loss function in these distribution fitting problems is the energy score, defined by
\begin{equation}
  L(\mathcal{X},\,\mathcal{O})
  =
  \dfrac{1}{M_{x}} \dfrac{1}{M_{o}}
  \sum\limits_{k=1}^{M_{x}} \sum\limits_{l=1}^{M_{o}}
  \lVert \rvec_k-\vec{o}_l \rVert_2
  -
  \frac{1}{2} \dfrac{1}{M_{x}} \dfrac{1}{M_{x}-1}
  \sum\limits_{k=1}^{M_{x}} \sum\limits_{l=1}^{M_{x}}
  \lVert \rvec_k-\rvec_{l} \rVert_2
  \label{eqn:energy_loss_empirical}
  \,,
\end{equation}
where $\mathcal{X}=\left\{\rvec_1, \rvec_2, \ldots, \rvec_{M_x}\right\}$ and $\mathcal{O}=\{\vec{o}_1, \vec{o}_2, \ldots, \vec{o}_{M_o}\}$ are sets of realizations from two distributions.
This loss function quantifies the distance between two distributions using only their realizations, without requiring explicit knowledge of the underlying distributions.

In distribution fitting, one draws samples $\mathcal{X}$ from a distribution $\density(\rvec;\, \parvec)$ parameterized by $\parvec$, while $\mathcal{O}$ denotes realizations from an unknown distribution---e.g., measurement data from physics experiments.
Distribution fitting is then formulated as an optimization problem of the form
\begin{equation}\label{eqn:dist_fitting_optimization}
    \parvec\opt \in \argmin_{\parvec\in\support{\parvec}} {L(\mathcal{X},\,\mathcal{O})} \,.
\end{equation}
By minimizing the energy score $L$, one aims to infer an optimal parameter $\parvec\opt$ such that $\density(\rvec;\, \parvec)$ closely approximates the unknown distribution underlying the observed data $\mathcal{O}$.

Gradient-based optimization procedures are typically employed to solve \eqref{eqn:dist_fitting_optimization}, which requires evaluating the gradient $\nabla_{\parvec} L$.
The dependence of $L$ on $\parvec$, however, is not explicit.
Specifically, $L$ depends indirectly on the distribution parameters through the realizations in $\mathcal{X}$ since their generation depends on the distribution parameter $\parvec$.
Thus, the calculation of $\nabla_{\parvec} L$ requires applying the chain rule as follows:
\begin{subequations}
\begin{equation}
  \nabla_{\parvec} L(\mathcal{X},\,\mathcal{O})
    =
      \sum\limits_{k=1}^{M_{x}}
      \left(\nabla_{\parvec} \rvec \right)_{\rvec=\rvec_k}^\mathsf{T}
      \,
      \nabla_{\rvec_k} L
      \,,
  \label{eqn:energy_loss_empirical_grad}
\end{equation}
where $\nabla_{\rvec_k} L$ has a closed form,
\begin{equation}
  \nabla_{\rvec_k} L
  =
  \frac{1}{M_x\, M_o} 
  \sum\limits_{l=1}^{M_o}
  \dfrac{\rvec_k - \vec{o}_l}{\lVert \rvec_k - \vec{o}_l \rVert_2}
  -
  \frac{1}{M_x\left(M_x-1\right)}  
  \sum\limits_{l=1}^{M_x}
  \dfrac{\rvec_k - \rvec_l}{\lVert \rvec_k - \rvec_l \rVert_2}
  \,,
\end{equation}
\end{subequations}
and $\nabla_{\parvec}\rvec$ is the space-parameter sensitivity (gradient of $\rvec$), which can be computed by using our proposed numerical algorithms (\OneDAlg, \FullInv, \InterpFull, \DiagApprox, and \InterpDiag).

Note that the energy score $L$ given by \eqref{eqn:energy_loss_empirical} is an empirical form (i.e., a Monte Carlo approximation) of the following continuous expectation:
\begin{equation}
  L
  =
  \dfrac{1}{M_o}
  \int\limits_{\rvec \in \domain{\density}}
    \sum\limits_{l=1}^{M_o}
    \lVert \rvec-\vec{o}_l \rVert_2 \,
  \density(\rvec)
  \diff\rvec
  -
  \frac{1}{2}
  \iint\limits_{\rvec, \rvec^{\prime} \in \domain{\density}}
    \lVert \rvec-\rvec^{\prime} \rVert_2 \,
    f(\rvec) f(\rvec^{\prime})
  \diff\rvec^{\prime}
  \diff\rvec
  \,,
  \label{eqn:energy_loss_continuous}
\end{equation}
where $\density(\rvec)\equiv \density(\rvec;\, \parvec)$.
This continuous form \eqref{eqn:energy_loss_continuous} is useful because it allows for analytical differentiation, providing a benchmark against which the numerical results obtained from the empirical approximation \eqref{eqn:energy_loss_empirical_grad} can be compared.
Specifically, in \eqref{eqn:energy_loss_continuous}, the parameter $\parvec$ influences the PDF $\density(\rvec;\, \parvec)$ explicitly.
This explicit dependence allows direct analytical differentiation with respect to $\parvec$:
\begin{equation}
  \nabla_{\parvec} L
  =
  \dfrac{1}{M_o}
  \int\limits_{\rvec \in \domain{\density}}
      \sum\limits_{l=1}^{M_o}
      \lVert \rvec-\vec{o}_l \rVert_2 \,
  \nabla_{\parvec} \density(\rvec)
  \diff\rvec
  -
  \iint\limits_{\rvec, \rvec^{\prime} \in \domain{\density}}
      \lVert \rvec-\rvec^{\prime} \rVert_2 \,
      f(\rvec)\nabla_{\parvec} f(\rvec^{\prime})
  \diff\rvec^{\prime}
  \diff\rvec
  \,.
  \label{eqn:energy_loss_continuous_grad}
\end{equation}
Note that if $\density$ is not normalized, then $\nabla_{\parvec} L$ is slightly different and is more complicated than \eqref{eqn:energy_loss_continuous_grad} because the normalization factor is itself a function of the distribution parameter $\parvec$.

As highlighted above, to validate whether the proposed formulae \eqref{eqn:1d_sensitivity_def}, \eqref{eqn:nd_sensitivity_def}, and \eqref{eqn:nd_sensitivity_approx} yield proper gradients for applications, we apply our numerical algorithms to the $\nabla_{\parvec} \rvec$ term in \eqref{eqn:energy_loss_empirical_grad} and then examine whether the resulting numerical $\nabla_{\parvec} L$ converges to its analytical form given by \eqref{eqn:energy_loss_continuous_grad} as $M_x$ increases.
The convergence rate is expected to follow that of Monte Carlo integration $\mathcal{O}(1/\sqrt{M_x})$ in the case of good numerical computations.
Once the calculation of $\nabla_{\parvec} L$ is validated, we continue by using it in the optimization process that solves \eqref{eqn:dist_fitting_optimization} and compare the resulting $\parvec\opt$ against the ground truth.

In our tests, whenever samples are required, we use a simple rejection sampling with a piecewise-constant proposal to draw samples from the target distribution.
Note that our proposed formulae and numerical algorithms are independent of the sampling procedure.
Specifically, our method always returns a deterministic $\nabla_{\parvec} \rvec$ for a given $\rvec$ and $\parvec$, regardless of how the samples were generated.
We discuss the results for 1-D beta distribution in \Cref{subsubsec:1d_beta_dist}, followed by the results for a 2-D proxy distribution in \Cref{subsubsec:2d_proxy}.


\subsubsection{Distribution Fitting with 1-D Beta Distribution.}
\label{subsubsec:1d_beta_dist}

Consider the 1-D beta distribution:
\begin{equation}
  \density(x; \parvec)
  =
  \frac{
    x^{\theta_1-1}(1-x)^{\theta_2-1}\mathrm{\Gamma}(\theta_1+\theta_2)
  }{
    \mathrm{\Gamma}(\theta_1)\mathrm{\Gamma}(\theta_2)
  }
  \,,
\end{equation}
where $\parvec\coloneqq\left[\theta_1, \theta_2\right]\tran$, $\theta_1 > 0,\, \theta_2 > 0$, $\mathrm{\Gamma}(\cdot)$ is the gamma function, and $x \in (0, 1)$.
The gradient of this PDF with respect to $\parvec$ is given by
\begin{equation}
  \nabla_{\parvec} \density(x; \parvec)
  =
  \begin{bmatrix}
    \pderiv{\density}{\theta_1} \\
    \pderiv{\density}{\theta_2}
  \end{bmatrix}
  =
  f(x; \parvec)
  \begin{bmatrix}
    \ln(x) + \psi(\theta_1+\theta_2) - \psi(\theta_1) \\
    \ln(1-x) + \psi(\theta_1+\theta_2) - \psi(\theta_2)
  \end{bmatrix}
  \,,
\end{equation}
where $\psi(\cdot)$ is the digamma function.
To define the loss function $L$, we sample observations $\mathcal{O}$  at $\theta_1=2.31$ and $\theta_2=1.627$ with $M_o=\num{10000}$.
The samples $\mathcal{X}$ are generated by setting $\theta_1=3$ and $\theta_2=1.4$.
The parameters at which we generate $\mathcal{X}$ are also used to calculate the analytical gradient \eqref{eqn:energy_loss_continuous_grad}.
The integrals in \eqref{eqn:energy_loss_continuous_grad} are numerically evaluated by using a fixed Gaussian--Legendre quadrature rule with $16$ points in both $\rvar$ and $\rvar^\prime$.

As an extra comparison, we also calculate \eqref{eqn:energy_loss_empirical_grad} numerically using the finite difference:
\begin{equation}
  \nabla_{\parvec} L = \sum_{i=1}^{2} \pderiv{L}{\param_i} \vec{e}_i \,;
  \qquad
  \pderiv{L}{\param_i}
  \approx
  \frac{
    L(\mathcal{X}^{+},\,\mathcal{O}) - L(\mathcal{X}^{-},\,\mathcal{O})
  }{
    2\Delta
  }
  \,,
  \label{eqn:finite_diff}
\end{equation}
where $\vec{e}_i$ is the $i$th unit vector in $\mathbb{R}^2$ and $\Delta > 0$ is the finite difference step size.
Here, the two sets of samples $\mathcal{X}^{+}$ and $\mathcal{X}^{-}$ are randomly sampled from perturbed distributions $\density(\rvec;\,\parvec+\Delta\cdot\vec{e}_i)$ and $\density(\rvec;\,\parvec-\Delta\cdot\vec{e}_i)$, respectively.
In our tests we use two different step sizes, $10^{-3}$ and $10^{-5}$.
Because of the stochastic nature of the random sampling procedure, the finite difference \eqref{eqn:finite_diff} is expected to produce noisy gradient estimations and may lead to divergence when employed in gradient-based optimization for solving \eqref{eqn:dist_fitting_optimization}.

\begin{figure}[htbp!]
  \centering
  \includegraphics[width=\linewidth]{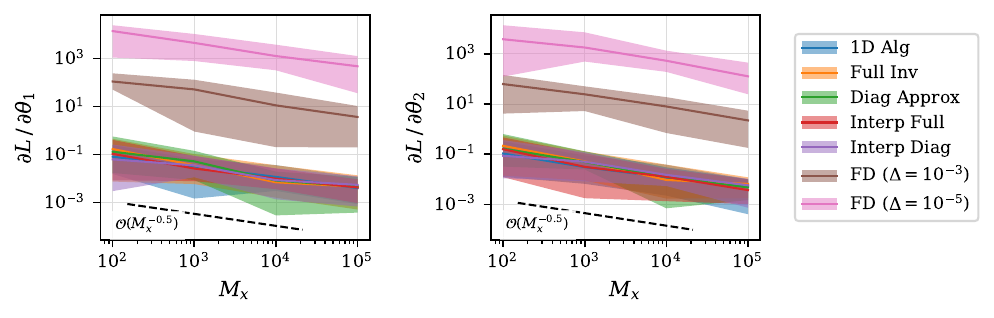}
  \caption{%
    Convergence of the relative errors of $\pderivi{L}{\theta_1}$ and $\pderivi{L}{\theta_2}$ with respect to $M_x$ (defined in \eqref{eqn:energy_loss_empirical}).
    FD denotes the naive finite difference in \eqref{eqn:finite_diff}.
    The figure's $y$-axis uses a logarithmic scale, so a nearly constant width of an uncertainty band signals decreasing uncertainty.
  }
  \label{fig:beta_grad_converge}
\end{figure}

\Cref{fig:beta_grad_converge} shows the error convergence in $\pderivi{L}{\theta_1}$ and $\pderivi{L}{\theta_2}$ with respect to $M_x$.
We observe the anticipated convergence rate of $\mathcal{O}(1\fracslash\sqrt{M_x})$, and all algorithms exhibit decreasing errors and uncertainty with increasing $M_x$.
Our algorithms also consistently deliver higher accuracy with reduced uncertainty compared with the naive finite difference method.
All proposed algorithms exhibit similar accuracy and uncertainty levels, as expected, since both \eqref{eqn:nd_sensitivity_def} and \eqref{eqn:nd_sensitivity_approx} reduce to \eqref{eqn:1d_sensitivity_def} for 1-D distributions.
In contrast, finite difference gradients demonstrate significantly lower accuracy and substantially higher uncertainty.

Interestingly, for the finite difference results, the larger step size produces both smaller errors and lower uncertainty compared with the smaller step size.
This counterintuitive behavior indicates that the naive finite difference method violates the consistency and convergence principles that are fundamental to finite difference schemes.
This suggests that the naive finite difference approximation \eqref{eqn:finite_diff} fails to provide a valid discrete approximation of $\nabla_{\parvec} L$.

\begin{figure}[htbp!]
  \centering
  \includegraphics[width=0.5\linewidth]{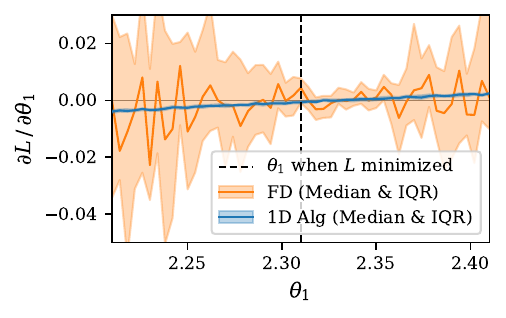}%
  \includegraphics[width=0.5\linewidth]{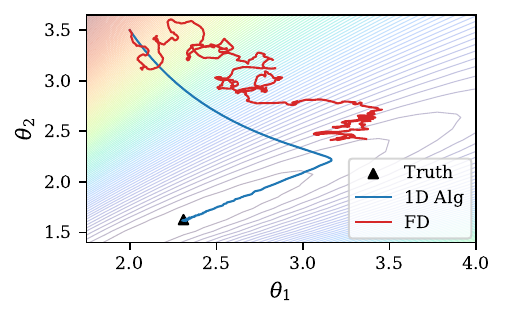}
  \caption{%
    Influence of numerical gradients using \OneDAlg and naive finite difference on gradient-based optimization.
    (Left) $\pderivi{L}{\theta_1}$ versus $\theta_1$ for a 1-D beta distribution with $\theta_2$ fixed at its optimal value.
    (Right) Training trajectories during optimization.
    The background contour shows the loss at different $(\theta_1, \theta_2)$.
    The blue and red lines show the loss trajectories during the iterative optimization using the ADAM optimizer.
  }
  \label{fig:beta_grad}
\end{figure}

\Cref{fig:beta_grad} visualizes how the incorrect gradients from the finite difference \eqref{eqn:finite_diff} fail the optimization process, and, in contrast, how our proposed algorithms help the optimization by providing effective and proper gradients.

The left plot in \Cref{fig:beta_grad} compares the computed $\pderivi{L}{\theta_1}$ using \OneDAlg and the naive finite difference at various values of $\theta_1$, with $\theta_2$ fixed at $1.627$ (the ground truth).
We used a training set $\mathcal{O}$ containing $M_o = \num{10000}$ samples and fixed $M_x$ at $\num{10000}$.
For each $\theta_1$ value, we repeated the gradient calculation $20$ times by sampling different $\mathcal{X}$ sets, allowing us to obtain the median and interquartile range (IQR).
We set the finite difference step size to $10^{-3}$.
Although we ran this calculation only with \OneDAlg, other algorithms are expected to perform similarly.
Since $\theta_2$ was already optimal, $\pderivi{L}{\theta_1}$ should ideally be zero at $\theta_1 = 2.31$ (the optimal value), although in practice the finite dataset sizes (i.e., $M_o$ in \eqref{eqn:energy_loss_empirical} and \eqref{eqn:energy_loss_continuous}) cause the optimal $\theta_1$ to deviate slightly from $2.31$.
The results demonstrate that our numerical algorithm yields nearly deterministic gradients and indicate a zero root slightly greater than $2.3$, which aligns with our expectation.
In contrast, the finite difference method produces highly stochastic results with multiple zero roots, a behavior that may lead the optimizer to incorrectly conclude that it has reached a minimum during training.

The right plot in \Cref{fig:beta_grad} shows the training trajectories of the optimization process using \OneDAlg and the naive finite difference method.
The ADAM optimizer was used with a fixed learning rate of $0.01$ and $\num{3000}$ training epochs.
As expected, \OneDAlg converges smoothly to the optimal $\theta_1$ and $\theta_2$.
In contrast, the naive finite difference fails to converge and exhibits erratic behavior.

In summary, the 1-D beta distribution case discussed here validates both the correctness and the effectiveness of the proposed method and the implementations offered by \DistroSA.

\subsubsection{Distribution Fitting with 2-D Proxy Distribution.}
\label{subsubsec:2d_proxy}

The second validation case considers the following unnormalized PDF parameterized by $5$ parameters:
\begin{equation}
  \density(\rvec; \parvec)
  =
  \rvar_1^{\param_1} (1 - \rvar_1)^{\param_2} \,
  \rvar_2^{\param_3} (1 - \rvar_2)^{\param_4} \,
  (1 + \param_5 \rvar_1 \rvar_2)
  \,,
  \label{eqn:2d_proxy_pdf}
\end{equation}
with $\rvec = \left[\rvar_1, \rvar_2\right]\tran \in \left(0, 1\right)^2$ and a parameter vector $\parvec = \left[\param_1, \param_2, \param_3, \param_4, \param_5\right]\tran$, where $-0.5 \le\! \param_1\! \le 1$, $2.75 \le\! \param_2 \le\! 4$, $0 \le\! \param_3 \le\! 1.3$, $3 \le\! \param_4 \le\! 4.5$, and $0 \le\! \param_5 \le\! 1.5$.
In nuclear physics, this distribution serves as a surrogate for more complex distributions in real-world applications, such as the one discussed in \Cref{subsec:app_dis}.

The ground truth parameters for generating observations $\mathcal{O}$ are $\parvec=\left[0.5, 3.0, 0.3, 4.0, 0.75\right]\tran$, with $M_o = \num{10000}$.
The $\parvec$ at which $\mathcal{X}$ is generated and both $L$ and $\nabla_{\parvec} L$ are evaluated is $\parvec=\left[0.25, 3.375, 0.65, 3.75, 0.1\right]\tran$.
The sample size $M_x$ is varied to investigate the convergence of the error.
The background grid for approximating conditionals $\cdf_i$ and their derivatives is a $1024 \times 1024$ rectilinear grid that is uniform on a logarithmic scale.
The integrals in \eqref{eqn:energy_loss_continuous} and \eqref{eqn:energy_loss_continuous_grad} are computed numerically using piecewise integration on a rectilinear grid with $50 \times 50$ nonuniform cells, where each cell is integrated using $9 \times 9$ Gaussian--Legendre quadrature points.
The cells are distributed logarithmically to capture high-density regions in the lower-left corner of the domain.

\begin{figure}[htbp!]
  \centering
  \includegraphics[width=\linewidth]{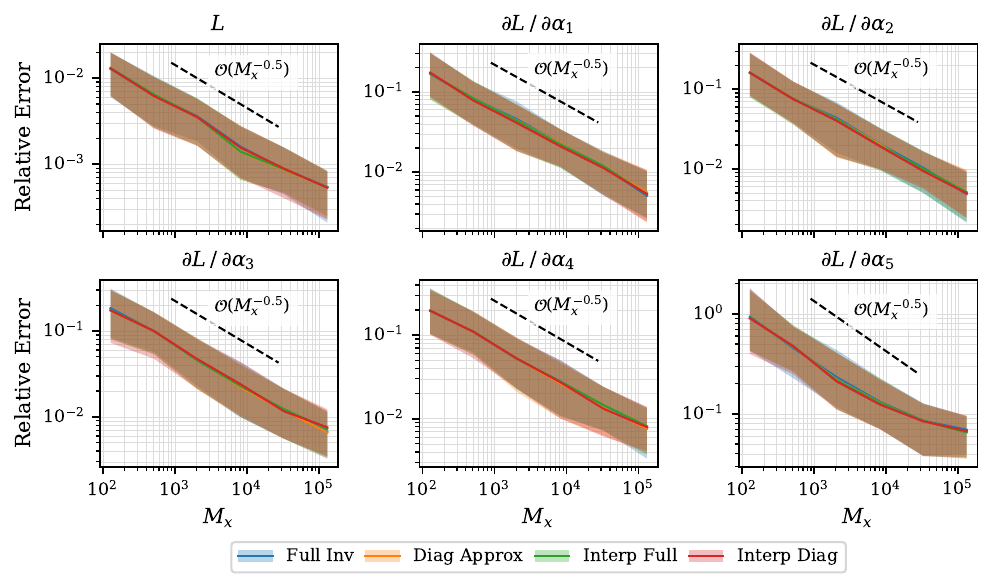}
  \caption{%
    Convergence of relative errors with respect to $M_\rvar$.
    The statistics, including the interquartile range represented by colored bands and the median shown as solid lines, are estimated from $100$ independent runs for a given $M_\rvar$.
    In each run, a new set of $\mathcal{X}$ is sampled, while the training data $\mathcal{O}$ remains unchanged.
    At a given $M_x$ and for a specific run, different algorithms use the same set of $\mathcal{X}$; thus, the plot of $L$ is expected to show identical behavior for all algorithms, since the loss calculation is independent of the algorithms in \DistroSA.
  }
  \label{fig:2d_proxy_converge}
\end{figure}

\Cref{fig:2d_proxy_converge} shows the error convergence of $L$ and the $\nabla_{\parvec} L$ from our numerical algorithms.
As observed in the 1-D beta distribution test in \Cref{subsubsec:1d_beta_dist}, the convergence rate is roughly $\mathcal{O}(1\fracslash\sqrt{M_\rvar})$, aligning with that of the Monte Carlo integration and thus our expectations.
All the algorithms behave similarly despite the different levels of approximation among them.

Next, we perform sample-based inference using the PDF \eqref{eqn:2d_proxy_pdf} with the same ground-truth parameters as previously described.
To accelerate computations, we used a coarser background grid resolution of $64\times 64$.
Despite this reduced resolution, our proposed algorithms still achieve good inference results, as demonstrated below.

We examine the convergence behavior of the inferred parameters using two different training datasets, $\mathcal{O}$ with $M_o = \num{10000}$ and $M_o = \num{50000}$ observations.
Since finite-sized training data cannot fully represent the continuous spatial domain of $\rvec$, we apply a bootstrapping analysis to quantify the uncertainty introduced by this limitation.
For each algorithm, we generate $100$ bootstrap samples by drawing observations from the original $\mathcal{O}$ with replacement.
To mitigate the risk of local minima, we repeat the inference process of each bootstrap sample $10$ times with different initialization points in $\parvec$-space.
We use the L-BFGS optimization algorithm for the inference process~\cite{liu_limited_1989}.
For a given bootstrap sample, the final inferred parameters are obtained from the repeat that gives the lowest final loss.

\begin{figure}[htbp!]
  \centering
  \includegraphics[width=\linewidth]{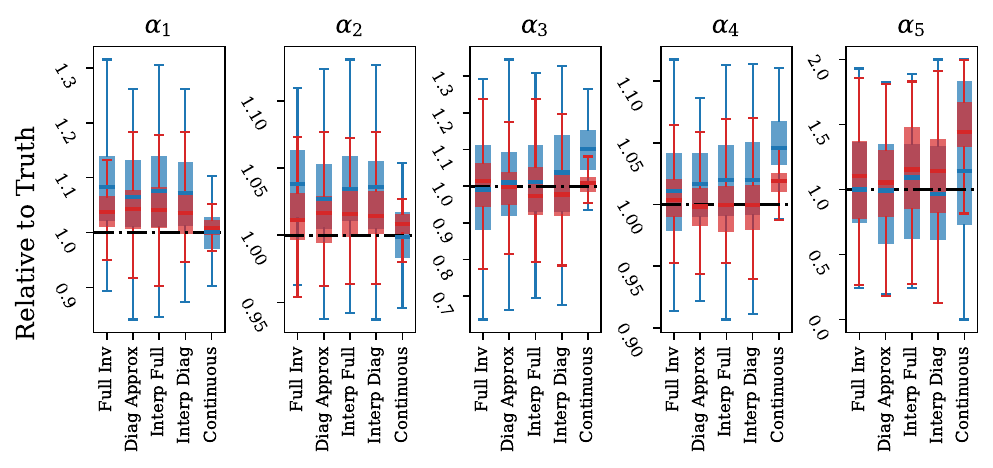}
  \caption{%
    Box plots for inferred parameters relative to the ground truth parameters. %
    Blue boxes are from $M_o=\num{10000}$, and red boxes are from $M_o=\num{50000}$. %
    The closer to $1.0$ (black dashed lines), the more accurate the result is. %
    Each box represents the distributions of inferred parameters from $100$ bootstrap samples.%
  }
  \label{fig:2d_proxy_trained_params}
\end{figure}

\begin{figure}[htbp!]
  \centering
  \includegraphics[width=0.5\linewidth]{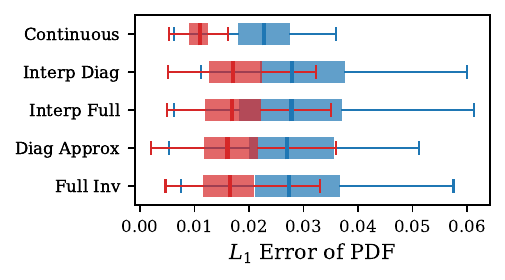}
  \caption{
    Box plot for $L_1$ error of the inferred PDF.
    The $L_1$ error is defined as $\int_{\domain{\density}} \lvert \density(\rvec; \hat{\parvec}) - \density(\rvec; \tilde{\parvec}) \rvert \diff \rvec$, where $\hat{\parvec}$ and $\tilde{\parvec}$ are inferred and true parameters, respectively.
    The blue boxes are from $M_o=\num{10000}$, and the red ones are from $M_o=\num{50000}$.
    Each box represents the distributions of $L_1$ errors using the inferred parameters from $100$ bootstrap samples.
  }
  \label{fig:2d_proxy_trained_pdf_errs}
\end{figure}

\Cref{fig:2d_proxy_trained_params} presents the interquartile range of the inferred parameters using each algorithm, as well as those obtained using the continuous energy score and its gradients \eqref{eqn:energy_loss_continuous} and \eqref{eqn:energy_loss_continuous_grad} for training.
\Cref{fig:2d_proxy_trained_pdf_errs} shows the IQR of $L_1$ errors of the inferred PDF.
These results show that as the number of training samples increased from $M_o = \num{10000}$ to $M_o = \num{50000}$, the accuracy improved and uncertainty became smaller in the inferred parameters across all algorithms.
This behavior was expected and supports the effectiveness of our proposed formulae and numerical algorithms.

The results also show that using the continuous loss and its gradients yields the smallest uncertainty and best accuracy.
In contrast, our numerical algorithms exhibit lower accuracy and higher uncertainty.
The results are consistent with the fact that only $\mathcal{O}$ behaves stochastically in the continuous energy score, while in the empirical energy score, $\mathcal{X}$ also behaves stochastically in addition to $\mathcal{O}$.
Despite these approximations, the use of the diagonal approximation and interpolation does not significantly worsen accuracy.

\begin{table}[htbp!]
  \centering
  \begin{tabular}{lccccc}
    \hline
    & {\footnotesize \FullInv}
    & {\footnotesize \DiagApprox}
    & {\footnotesize \InterpFull}
    & {\footnotesize \InterpDiag}
    & {\footnotesize \texttt{Continuous}}
    \\
    \hline
    {\small $M_o=10$k} & $21 \pm 19$ & $20 \pm 18$ & $21 \pm 20$ & $19 \pm 18$ & $3 \pm 1$ \\
    {\small $M_o=50$k} & $39 \pm 35$ & $35 \pm 32$ & $36 \pm 32$ & $34 \pm 33$ & $4 \pm 1$ \\
    \hline
  \end{tabular}
  \caption{%
    Computational time (in seconds) required for inference using different algorithms. %
    The values shown are the averages and standard deviations of $100$ bootstraps and $10$ repeated runs per bootstrap. %
    The computations were performed on a single NVIDIA A100 GPU (40GB variant). %
  }
  \label{tab:algorithm_comparison}
\end{table}

\Cref{tab:algorithm_comparison} presents the computational time required for inference across different algorithms.
Although \InterpDiag demonstrates marginally superior computational performance, the observed improvement falls short of our initial expectations.
We believe that the primary bottleneck limiting computational efficiency stems from the evaluation of the empirical energy score, which has quadratic complexity, $\mathcal{O}(M_x M_o + M_x^2)$.
This bottleneck dominates the overall runtime, making the differences in computational costs less observable across different algorithms.
On the other hand, the continuous energy score formulations \eqref{eqn:energy_loss_continuous} and \eqref{eqn:energy_loss_continuous_grad} achieve notably faster times due to their fundamentally different computational structure.
In these cases, $M_x$ and $M_o$ represent the number of quadrature points used in numerical integration, which is significantly smaller than the corresponding sample sizes in the empirical energy score computation.
Consequently, future work should incorporate more granular performance profiling to understand the true computational efficiency of different algorithms.
Alternatively, replacing the energy score with another stochastic loss with subquadratic complexity may also help better distinguish the performance differences among different algorithms.

The time statistics in \Cref{tab:algorithm_comparison} are computed from \num{1000} optimization runs ($100$ bootstraps with $10$ repeated runs each).
Bootstrap resampling changes the stochastic loss surface, while repeated runs start from different initial parameter values and may require different numbers of iterations to converge.
Together, these effects explain the large standard deviations observed in \Cref{tab:algorithm_comparison}.

\subsection{Application: Deep Inelastic Scattering}
\label{subsec:app_dis}

This section showcases the usefulness of our work in providing the space-parameter sensitivity required by the optimization process in a more realistic application of sample-based inference in nuclear physics.
This application aims to infer the parameters of quantum correlation functions (QCFs) via measurement data collected during deep inelastic scattering (DIS) experiments.
We refer readers to references \cite{chuang_characterization_2024, ellis_qcd_1996} for physical and mathematical details of this parameter inference problem.
Here we give only a high-level understanding of the application and the configurations of our execution.

A QCF describes the probability of finding a specific type of quark (up, down, etc.) or gluons in a hadron (e.g., a proton or neutron) carrying a certain momentum fraction $x$ of the parent hadron at a probing energy scale $Q^2$ measured during particle collision experiments.
QCFs are commonly denoted as, for example, $u(x, Q^2)$, $d(x, Q^2)$, and $g(x, Q^2)$ for up quark, down quark, and gluon, respectively.
In other words, these QCFs can be treated as unnormalized probability density functions or likelihood functions of the random vector $\rvec \coloneqq \left[x, Q^2\right]^\mathsf{T}$.

However, $\left[x, Q^2\right]^\mathsf{T}$ realizations from each individual QCF cannot be directly observed during experiments: although the DIS experiments observe the realizations of $\left[x, Q^2\right]^\mathsf{T}$ of a quark/gluon, they cannot determine which type of quark or gluon is broken away from the hadron.
In other words, what these experiments measure are actually the realizations of a collective distribution called the differential cross section, which combines contributions from all possible quark types and gluons that could have been involved in the collision.
This collective distribution is denoted as $f(x, Q^2)$ and depends on all individual QCFs: $f(x, Q^2) \coloneq f(x, Q^2, u(x, Q^2), d(x, Q^2), \ldots, g(x, Q^2))$.
Different hadron types have different differential cross sections; for example, $f_n$ and $f_p$ denote the differential cross sections of neutrons and protons.
By assuming some parametric forms for QCFs, for example, $u(x, Q^2;\, \parvec)$, $d(x, Q^2;\, \parvec)$, and $g(x, Q^2;\, \parvec)$, this inference application infers the parameter $\parvec$ by comparing the observed $\left[x, Q^2\right]^\mathsf{T}$ realizations against the samples drawn from $f_n$ and $f_p$.
The dependence of $f_n$ and $f_p$ on $\parvec$ is via QCFs.
$f_n$ and $f_p$ therefore serve as the target likelihood functions (or unnormalized probability density functions) in this sample-based inference.

One major challenge in this application is that $f_n$ and $f_p$ lack closed-form expressions.
They are expensive computer simulation routines that solve systems of integro-differential equations.
Therefore, we use only the \InterpDiag algorithm to estimate $\nabla_{\parvec} \rvec$ to save computational costs, since it requires fewer PDF evaluations than other algorithms.

We configure this inference application as follows.
We use pseudo-experimental data as the training data $\mathcal{O}$ rather than real-world experimental data.
The pseudo-experimental data are synthetically generated from the computer simulation routines of $f_n$ and $f_p$ by some given ground-truth parameters, allowing us to control the ground truth of the QCFs and establish baselines for comparison.
The loss function $L$ is the empirical energy score, as discussed in \Cref{subsec:validation}.
We use two sets of training data: one for protons (compared against $f_p$) and another for neutrons (compared against $f_n$), with the final loss being the sum of losses from both sets.
Both neutron data and proton data are used because $f_p$ and $f_n$ depend on the same QCFs.
We apply bootstrapping analysis to quantify uncertainty, similar to our approach in \Cref{subsubsec:2d_proxy}.
From the original training dataset, $\num{1000}$ bootstrap samples are generated, and $20$ repeated runs are performed for each bootstrap sample to mitigate the risk of local minima.
We test convergence of uncertainty and accuracy using training dataset sizes $M_o = \num{2000}$, $\num{5000}$, $\num{10000}$, and $\num{50000}$, with $M_x$ matching each $M_o$.
The parametric QCFs share the same parameter vector, defined as $\parvec = \left[g_{\alpha}, g_{\beta}, u_{\alpha}, u_{\beta}, d_{\alpha}, d_{\beta}\right]\tran$.
Refer to \cite{chuang_characterization_2024, ellis_qcd_1996} for the definitions of these parametric functions.

The domain of this inference problem is bounded by the physical constraints~\cite{ellis_qcd_1996} $m_c^2 \le Q^2 \le s - W_{\text{min}}^2$ and $\frac{Q^2}{s - M^2} \le x \le \frac{Q^2}{W_{\text{min}}^2 - M^2 + Q^2}$, where the charm quark mass is $m_c=1.28\,\text{GeV}$, the lepton-target center-of-mass energy squared is $s=140^2\, \text{GeV}^2$, the minimum hadronic final-state mass is $W_{\text{min}}=\sqrt{10}\,\text{GeV}$, and the target hadron mass is $M=0.93891897\,\text{GeV}$.
To construct the background grid, we transform $(x, Q^2)$ to $(x^\prime, Q^{2\prime}) \in (0, 1]^2$.
The grid is a logarithmically uniform $64 \times 64$ Cartesian grid in this transformed domain.
We use this resolution because it matches the inference runs in \Cref{subsubsec:2d_proxy}, where the proxy distribution serves as a domain-informed surrogate for the QCFs in the mapped unit-square domain.

\begin{figure}[htbp!]
  \centering
  \includegraphics[width=\linewidth]{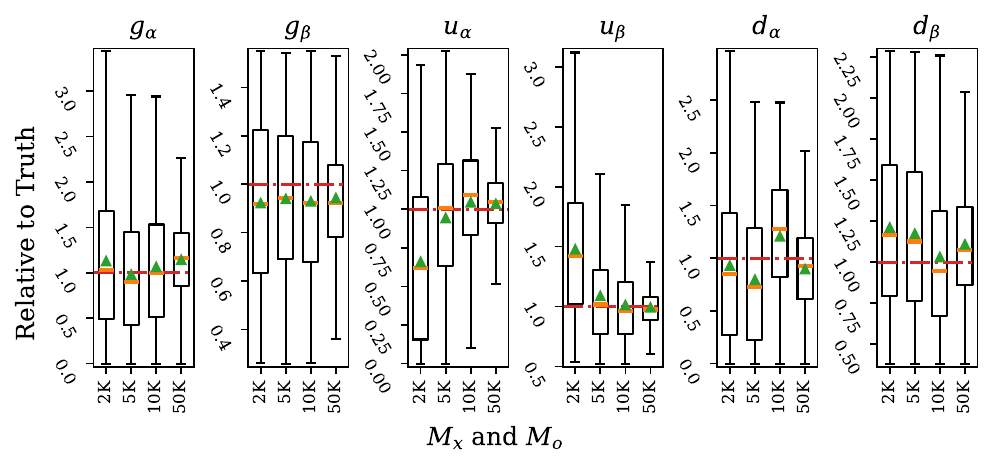}
  \caption{%
    Box plots of inferred parameters relative to ground truths for the DIS parameter inference application.
    Each box represents the distribution of inferred parameters from $\num{1000}$ bootstrap samples.
    The orange horizontal lines and green triangle markers represent the medians and means, respectively.
    Since the values shown are inferred parameters relative to the ground truths, the ground truths are always at $1.0$ (denoted by the horizontal red dashed lines) in each plot.
  }
  \label{fig:dis_parvec}
\end{figure}

\begin{figure}[htbp!]
  \centering
  \includegraphics[width=\linewidth]{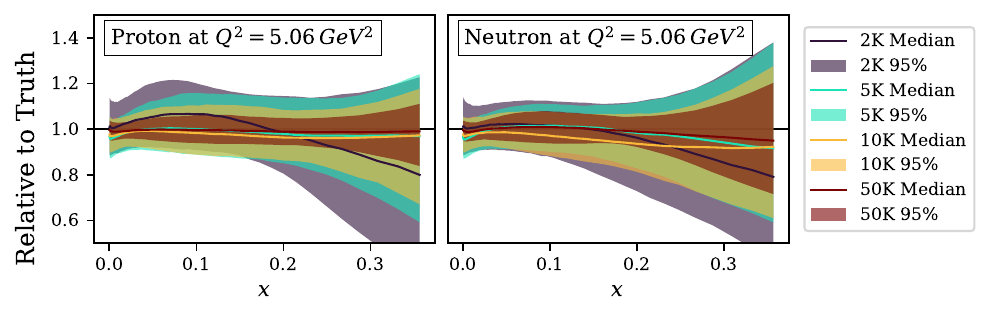}
  \caption{
    Inferred differential cross sections, $f_p(x, Q^2)$ (left) and $f_n(x, Q^2)$ (right), at $Q^2=5.06\,\mathrm{GeV}^2$.
    The values are shown relative to the ground truths.
    Therefore, the closer to $1.0$, the more accurate the result.
    The solid lines represent the medians of results using different $M_o$ (note that $M_x=M_o$).
    The colored bands denote the $95\%$ confidence intervals estimated from $\num{1000}$ bootstrap samples.
  }
  \label{fig:dis_xsec}
\end{figure}

\Cref{fig:dis_parvec} shows the inferred parameters $\parvec$.
As $M_o$ increases, the uncertainty of the inferred parameters generally decreases, as indicated by the shrinking box sizes in the plots.
However, the uncertainty of $d_{\beta}$ converges much more slowly than the other parameters.
The accuracy also generally converges to the ground truth, with means and medians approaching $1.0$ as $M_o$ increases.
The parameter $g_{\beta}$ is an exception, where the means and medians do not converge to the ground truth.
The slow convergence of $d_{\beta}$ and non-convergence of $g_{\beta}$ likely occur because these parameters affect the shape of $f_n$ and $f_p$ in the high-$x$ region, where the probability density is extremely small.
Without an extremely large $M_o$, these two parameters are difficult to identify correctly.
This behavior is confirmed by the uncertainty and accuracy of the inferred $f_n$ and $f_p$ shown in \Cref{fig:dis_xsec}.

\Cref{fig:dis_xsec} shows the inferred $f_n$ and $f_p$ obtained by using the inferred parameters.
The uncertainty of both functions decreases with increasing $M_o$, as evidenced by the shrinking colored bands.
Moreover, the accuracy of both functions converges toward the ground truths, as shown by the solid lines approaching $1.0$ with increasing $M_o$.
These results suggest that the slow convergence of $d_{\beta}$ and non-convergence of $g_{\beta}$ have minimal impact on $f_n$ and $f_p$, which are de facto PDFs.

\begin{figure}[htbp!]
  \centering
  \includegraphics{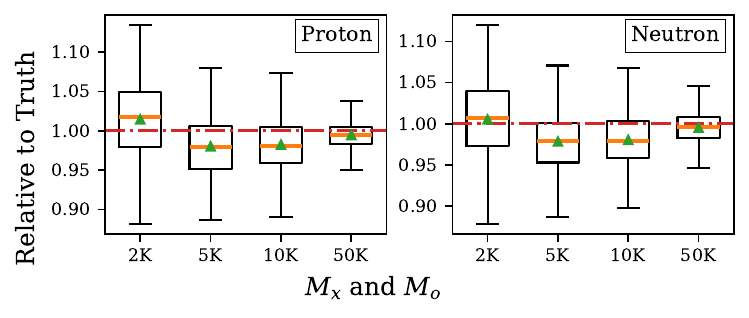}
  \caption{%
    Box plots of inferred total cross sections.
    The total cross section of a differential cross section, $f$, is defined as $\int_{\Omega_f} f \diff x \diff Q^2$.
    Each box represents the distribution of inferred total cross sections from $\num{1000}$ bootstrap samples.
    The orange horizontal lines and green triangle markers represent the medians and means, respectively.
    Since the values shown are relative to the ground truths, the ground truths are always at $1.0$ (denoted by the horizontal red dashed lines) in each plot.
  }
  \label{fig:dis_totxsec}
\end{figure}

\Cref{fig:dis_totxsec} shows the total cross sections computed from the inferred parameters.
The total cross section represents the normalization factor of a differential cross section, mathematically equivalent to its integral over the entire domain.
Because $f_n$ and $f_p$ are unnormalized, QCFs with different parameter values produce different normalization factors.
Therefore, if the inferred parameters are correct---meaning the reconstructed QCFs accurately represent the true distributions---the resulting normalization factor should match the ground truth value.
These normalization factors provide an independent validation metric for assessing the accuracy of the parameter inference.
The results demonstrate that uncertainty in the normalization factors decreases and accuracy converges toward the ground truth with increasing $M_o$, further confirming that $d_{\beta}$ and $g_{\beta}$ have minimal influence on the differential cross section.

\begin{figure}[htbp!]
  \centering
  \includegraphics{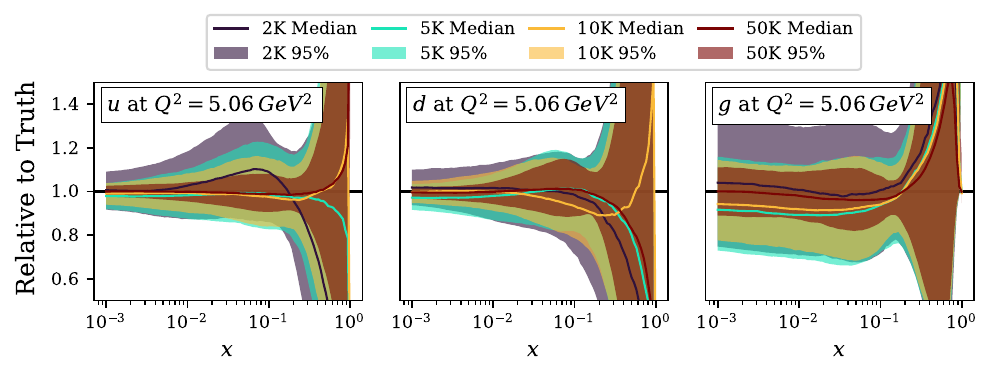}
  \caption{
    Inferred QCFs, $u(x, Q^2)$, $d(x, Q^2)$, and $g(x, Q^2)$, relative to ground truths at $Q^2=5.06\,\mathrm{GeV}^2$.
    The values are shown relative to the ground truths.
    Therefore, the closer to $1.0$, the more accurate the result.
    The solid lines represent the medians of results using different $M_o$ (note that $M_x=M_o$).
    The colored bands denote the $95\%$ confidence intervals estimated from $\num{1000}$ bootstrap samples.
  }
  \label{fig:dis_qcf}
\end{figure}

\Cref{fig:dis_qcf} shows the inferred QCFs, $u(x, Q^2)$, $d(x, Q^2)$, and $g(x, Q^2)$.
For low-$x$ regions, all QCFs behave as expected: uncertainty decreases and accuracy improves toward the ground truth as $M_o$ increases.
For $d(x, Q^2)$ and $g(x, Q^2)$, however, the high-$x$ regions show little response to changes in $M_o$.
Since the high-$x$ regions of $f_n$ and $f_p$ in \Cref{fig:dis_xsec} behave as expected, this suggests that the high-$x$ regions of these two QCFs have little effect on the differential cross sections, making it hard to identify the correct values of $d_{\beta}$ and $g_{\beta}$.
In contrast, for $u(x, Q^2)$, the high-$x$ region where changing $M_o$ has little effect occurs at larger $x$ values than the other two QCFs, which may explain why $u_{\beta}$ is easier to infer.

The primary objective of this nuclear physics application is to infer the QCFs, which cannot be directly observed in experiments.
The successful reconstruction shown in \Cref{fig:dis_qcf} demonstrates that our approach achieves this goal.
Since our focus here is to demonstrate the practical application of our proposed method, we have omitted many technical details of this inference problem.
Interested readers can find more details in~\cite{chuang_characterization_2024, ellis_qcd_1996}.


\section{Discussion and Concluding Remarks}
\label{sec:discussion}

This work presents three mathematical formulations (\Cref{sec:methods}) for computing sensitivities of random vectors with respect to distribution parameters: \eqref{eqn:1d_sensitivity_def} for 1-D distributions, \eqref{eqn:nd_sensitivity_def} for distributions of arbitrary dimensionality with smooth PDFs, and \eqref{eqn:nd_sensitivity_approx} also for distributions of arbitrary dimensionality but with the diagonal approximation.
The last one has lower accuracy but lower computational cost.
Since these formulae rarely yield closed-form expressions for most probability distributions, this work also provides five second-order-accurate numerical algorithms to approximate them.
All algorithms are implemented in our open-source package \DistroSA (Distributional Sensitivity Analysis)~\cite{attia_distrosa_2025, attia_distrosa_2025-1}.

The key advantage of our method is its independence from sampling procedures, making it particularly valuable when reparameterization is infeasible---such as in sample-based inference problems where both sampling and PDF evaluation are embedded within complex or black-box forward simulations of physical phenomena.

In our numerical experiments, we first verified that our numerical algorithms accurately approximate the proposed formulae by comparing the numerical solutions against closed-form solutions for 1-D and 2-D Gaussian distributions.
We then validated the effectiveness of the proposed formulae in calculating the gradients of the energy score and in sample-based inference.
Additionally, our nuclear physics benchmark demonstrated that even the \InterpDiag algorithm, despite having the highest degree of approximation, performed well when the underlying distribution is non-analytical and embedded within computationally intensive forward simulators.

From the perspective of accuracy, numerical algorithms with higher degrees of approximation (\InterpFull---interpolation, \DiagApprox---diagonal approximation, and \InterpDiag---diagonal approximation plus interpolation) perform similarly to \FullInv---which incorporates the fewest approximations.
This suggests that these computationally cheaper algorithms may be advantageous in computationally intensive applications.
However, a caveat of our experiments is that although algorithms with more approximations were indeed less computationally expensive, the performance gains were smaller than expected.
We attribute this result to the energy score's quadratic computational complexity, which dominates the total runtime.
To properly assess the computational efficiency of each \DistroSA algorithm, future work should isolate the cost of loss-function evaluation through detailed profiling and performance analysis.

In practical applications, the computational domain is usually determined by physical constraints when the underlying distribution has bounded support.
For unbounded distributions, domain-edge effects arise because a finite computational domain is used to approximate an infinite-support distribution.
As illustrated in \Cref{subsec:verification}, this domain-edge effect appears as a persistent bias in the CDF gradients near the computational-domain boundaries, and this bias cannot be removed by refining the grid alone.
The practical criterion is therefore to choose a computational domain that is larger than the domain of interest, with enough padding that boundary-biased CDF gradients do not affect the quantities of interest.

A direct way to choose the domain size is to repeat the calculation on increasingly large computational domains and verify that the quantities of interest stabilize within the domain of interest.
If this is too expensive, one can first estimate an initial domain using a simple empirical surrogate fitted to the training data.
For example, one can estimate statistical moments from the training data, use closed-form formulas to fit a simple distribution such as a Gaussian, gamma, or Student-$t$ distribution, and then choose domain bounds so that the surrogate tail probability outside each bound is below a prescribed tolerance.
For multivariate problems, because this surrogate is used only to estimate the domain size, this tail-probability criterion can be applied independently in each spatial direction without modeling correlations between components.

Grid spacing can be selected similarly by refining the background grid until the quantities of interest are insensitive to further refinement.
To reduce the cost, calculations on at least three grid resolutions can be used with techniques such as three-grid Richardson extrapolation~\cite{roache_perspective_1994} to estimate the discretization error and the converged values.

These convergence checks on the quantities of interest are necessary because domain-edge effects and background-grid resolution may affect gradient calculations without necessarily degrading the final results.
For example, even if the computed gradients contain errors throughout the domain, a stochastic gradient-based optimizer may still tolerate these errors and converge to accurate model parameters, which are the quantities of interest in this inference problem.

The current numerical algorithms, while promising, are limited to low-dimensional cases and use relatively basic numerical techniques such as finite differences and multilinear interpolation.
These constraints point toward several promising directions for future research.
First, benchmarking in higher-dimensional settings (i.e., dimensionality $>2$) will help assess scalability.
Second, more advanced numerical schemes could be explored to improve either accuracy or efficiency---particularly in higher dimensions.
For instance, h/p-refined piecewise polynomials or mesh-free interpolation schemes could replace the current combination of finite differences and multilinear interpolation for both function approximation and derivative computation.
Alternatively, integrating \DistroSA with modern machine learning frameworks such as PyTorch would enable computation of conditional CDF derivatives via automatic differentiation.

The Rosenblatt transform also motivates future algorithm design.
We did not count PDF evaluations for a Rosenblatt-based variant in \Cref{subsec:comp_cost} because direct substitution into the present algorithms would not give a fair comparison.
For example, replacing $\density_i$ in \Cref{alg:get_sensitivity_nd} with nested trapezoidal rules for \eqref{eqn:Rosenblatt_Transform_3} would require $\mathcal{O}(\sum_{i=1}^{N}\prod_{j=i}^{N} K_j)$ PDF evaluations for $\pseudocdf$ at one spatial point, compared with $\mathcal{O}(\sum_{i=1}^{N} K_i)$ in \Cref{alg:get_sensitivity_nd}.
A dedicated implementation could reduce the PDF-evaluation cost by reusing and interpolating joint-PDF values at grid vertices.
It could also exploit the triangular structure of the Rosenblatt transform, i.e., $\pderivi{\cdf_i}{\rvar_j}=0$ for $j > i$, which eliminates some finite-difference derivative terms and their perturbed PDF evaluations.
However, the marginal-conditional integrations in \eqref{eqn:Rosenblatt_Transform_3} would introduce additional arithmetic cost.
In contrast, the triangular structure may reduce arithmetic cost because triangular matrix inversion is cheaper than the full matrix inversion used here.
A concrete algorithm and implementation are needed to assess the net effect of these competing factors.
We therefore leave the design, implementation, and V\&V of a Rosenblatt-specific algorithm for future work.

In summary, this work introduces a practical method for computing sensitivities of random vectors with respect to distributional parameters, particularly in settings where no explicit reparameterization is available.
The presented verification, validation, and application benchmarks confirm both the correctness and usefulness of our method.
Future efforts will aim to enhance numerical efficiency and develop a deeper understanding of the numerical properties of the proposed algorithms.

  \section*{Acknowledgments and Funding Sources}
  This material is based upon work supported by the U.S. Department of Energy, Office of Science, Office of Advanced Scientific Computing Research (ASCR) and the Scientific Discovery through Advanced Computing (SciDAC) Program Nuclear Physics partnership titled ``Femtoscale Imaging of Nuclei using Exascale Platforms'' and the FASTMath Institute programs under Contract No. DE-AC02-06CH11357.%

  \bibliographystyle{elsarticle-num}
  \bibliography{references.bib}

  \clearpage

  \appendix


\section{Additional Algorithms}
\label[appendix]{sec:algorithms}

This appendix describes the auxiliary algorithms and equations required by previously described algorithms.

\Cref{alg:get_cdf_1d} evaluates CDF values at the vertices of a 1-D grid $\vec{v}$ for a given PDF $\density$.
The implementation employs the trapezoidal rule for numerical integration, providing second-order accuracy while maintaining robustness for arbitrary 1-D distributions and grid configurations.

\begin{algorithm}[htbp!]
  \SetKwFunction{getcdfoned}{get\_cdf\_1d}
  \caption{CDF evaluation at grid vertices for 1-D distributions}
  \label{alg:get_cdf_1d}

  \KwIn{1-D $\density$, $\vec{v}$, $\parvec$}

  \KwOut{
    $\vec{\Phi}$ (normalized CDF at vertices; $\vec{\Phi} \in \Rnum{K}$),
    $c$ (normalization factor)
  }

  \BlankLine
  \Fn{\getcdfoned{$\density$, $\vec{v}$, $\parvec$}}{

    $\vec{\Phi} \gets \mathbf{0}_{K}$\;

    $\density_{l} \gets \density(v^1,\,\parvec)$\;

    \For(\tcp*[h]{integrating from $v^1$ to $v^k$}){$k\gets 2$ \KwTo $K$}{

      $\density_{r} \gets \density(v^k,\,\parvec)$\;

      $\vec{\Phi}[k] \gets \vec{\Phi}[k-1] + (\density_l + \density_r)
        \times (v^k - v^{k-1}) \fracslash 2$\;

      $\density_l \gets \density_r$\;
    }

    $c \gets \vec{\Phi}[K]$\;

    $\vec{\Phi} \gets \vec{\Phi} \fracslash c$ \tcp*[l]{array scaling w/ a scalar}

    \KwRet ($\vec{\Phi}$, $c$)\;
  }
\end{algorithm}

\Cref{alg:get_conditionals}, used by \Cref{alg:get_sensitivity_nd_grid}, numerically computes all 1-D conditionals at all vertices using the trapezoidal rule while reusing the joint PDF values at those vertices.

\begin{algorithm}[htbp!]
  \SetKwFunction{getconditionals}{get\_conditionals}
  \caption{Calculator for all 1-D conditional PDFs/CDFs at N-D grid vertices}
  \label{alg:get_conditionals}

  \KwIn{$\density$, $\vec{V}$, $\parvec$}

  \KwOut{
    \hangindent=1em
    $\mat{\phi}$ (sequence of conditional PDFs on vertices, i.e.,
      $\mat{\phi} = (\mat{\phi}_1, \dots, \mat{\phi}_\nrvars)$,
      where $\mat{\phi}_i \in \Rnum{K_1 \times \dots \times K_\nrvars}$ $\forall i$),
    $\mat{\Phi}$ (sequence of conditional CDFs on vertices, i.e.,
      $\mat{\Phi} = (\mat{\Phi}_1, \dots, \mat{\Phi}_\nrvars)$,
      where $\mat{\Phi}_i \in \Rnum{K_1 \times \dots \times K_\nrvars}$ $\forall i$)
  }

  \BlankLine

  \Fn{\getconditionals{$\density$, $\mat{V}$, $\parvec$}}{

    \BlankLine

    \tcp{initialize data holders}

    $\mat{\phi} \gets (\mat{\phi}_1, \dots, \mat{\phi}_\nrvars)$
      where
      $\mat{\phi}_i \gets \mathbf{0}_{K_1 \times \dots \times K_\nrvars}, \forall i$\;

    $\mat{\Phi} \gets (\mat{\Phi}_1, \dots, \mat{\Phi}_\nrvars)$
      where
      $\mat{\Phi}_i \gets \mathbf{0}_{K_1 \times \dots \times K_\nrvars}, \forall i$\;
    $\mat{\eta} \gets
      \mathbf{0}_{K_1 \times \dots \times K_\nrvars}$\tcp*[l]{to hold joint PDF}

    \BlankLine

    \tcp{compute joint PDF values at all vertices; $\vec{k}=(k_1, \ldots, k_\nrvars)$}
    \ForEach(){%
      $\vec{k} \in \prod_{i=1}^{\nrvars} \{1, \ldots, K_i\}$%
    }{
      $\vec{s} \gets (v_{1}^{k_1}, v_{2}^{k_2}, \dots, v_{\nrvars}^{k_{\nrvars}})$\;
      $\mat{\eta}[\vec{k}] \gets \density(\vec{s},\,\parvec)$\;
    }

    \BlankLine
    \tcp{compute $i$-th conditional CDF at all vertices}
    \For(){$i \gets 1$ \KwTo $\nrvars$}{

      \ForEach(\tcp*[h]{$k_i$ must run in increasing order}){%
        $\vec{k} \in \prod_{i=1}^{\nrvars} \{1, \ldots, K_i\}$%
      }{

        \lIf(\tcp*[h]{skip the 1st vertex as it's zero}){$k_i = 1$}{continue}

        $\vec{k}^\prime \gets (k_1, \dots, k_{i}-1, \dots, k_\nrvars)$\;

        $\mat{\Phi}_{i}[\vec{k}] \gets
          \mat{\Phi}_{i}[\vec{k}^\prime] + (\mat{\eta}[\vec{k}^\prime] +
          \mat{\eta}[\vec{k}]) \times (v_{i}^{k_{i}} - v_{i}^{k_{i}-1})
          \fracslash
          2$\;
      }
    }

    \BlankLine
    \tcp{normalization}
    \For(){$i \gets 1$ \KwTo $\nrvars$}{
      \ForEach(\tcp*[h]{$k_i$ must run in increasing order}){%
        $\vec{k} \in \prod_{i=1}^{\nrvars} \{1, \ldots, K_i\}$%
      }{

        $\vec{k}^\prime \gets
          (k_1, \dots, k_{i-1}, K_i, k_{i+1}, \dots, k_\nrvars)$
          \tcp*[l]{the last vertex}

        $\mat{\phi}_{i}[\vec{k}] \gets
          \mat{\eta}[\vec{k}] \fracslash \mat{\Phi}_{i}[\vec{k}^\prime]$\;

        $\mat{\Phi}_{i}[\vec{k}] \gets
          \mat{\Phi}_{i}[\vec{k}] \fracslash \mat{\Phi}_{i}[\vec{k}^\prime]$\;
      }
    }

    \KwRet ($\mat{\phi}$, $\mat{\Phi}$)\;
  }
\end{algorithm}

The second-order finite-difference formulae used in \Cref{alg:get_sensitivity_nd_grid} are described below:
\begin{itemize}
  \item Second-order forward difference for first-order derivatives
    \begin{equation}
      \left.\pderiv{\phi}{x}\right|_{x=x_k}
      \approx
      \frac{
        -
        h_{k+1} \left(2 h_{k} + h_{k+1}\right) \phi_{k}
        +
        \left(h_{k} + h_{k+1}\right)^{2} \phi_{k+1}
        -
        h_{k}^{2} \phi_{k+2}
      }{
        h_{k} h_{k+1} \left(h_{k} + h_{k+1}\right)
      }
      \,,
      \label{eqn:forward_difference}
    \end{equation}
    where $h_{k} \coloneqq x_{k+1} - x_{k}$ and $h_{k+1} \coloneqq x_{k+2} - x_{k+1}$.
  \item Second-order backward difference for first-order derivatives
    \begin{equation}
      \left.\pderiv{\phi}{x}\right|_{x=x_k}
      \approx
      \frac{
        h_{k-2} \left(2 h_{k-1} + h_{k-2}\right) \phi_{k}
        -
        \left(h_{k-1} + h_{k-2}\right)^{2} \phi_{k-1}
        +
        h_{k-1}^{2} \phi_{k-2}
      }{
        h_{k-1} h_{k-2} \left(h_{k-1} + h_{k-2}\right)
      }
      \,,
      \label{eqn:backward_difference}
    \end{equation}
    where $h_{k-1} \coloneqq x_{k} - x_{k-1}$ and $h_{k-2} \coloneqq x_{k-1} - x_{k-2}$.
  \item Second-order central difference for first-order derivatives
    \begin{equation}
      \left.\pderiv{\phi}{x}\right|_{x=x_k}
      \approx
      \frac{
        -
        h_{k}^{2} \phi_{k-1}
        +
        \left(h_{k} - h_{k-1}\right) \left(h_{k} + h_{k-1}\right) \phi_{k}
        +
        h_{k-1}^{2} \phi_{k+1}
      }{
        h_{k} h_{k-1} \left(h_{k} + h_{k-1}\right)
      }
      \,,
      \label{eqn:central_difference}
    \end{equation}
    where $h_{k} \coloneqq x_{k+1} - x_{k}$ and $h_{k-1} \coloneqq x_{k} - x_{k-1}$.
\end{itemize}
These finite difference formulae maintain overall second-order accuracy for the second solution algorithm by ensuring consistent error convergence rates at both interior and boundary vertices.


\section{Analytical Jacobians for 1-D and 2-D Gaussian Distributions}
\label[appendix]{sec:analytical_gaussian}

In this section, we derive the closed-form expressions for \cref{eqn:1d_sensitivity_def,eqn:nd_sensitivity_def,eqn:nd_sensitivity_approx} for 1-D and 2-D Gaussian distributions.
These closed forms were used in the verification in \Cref{subsec:verification}.
    
\subsection{1-D Gaussian Distribution}
\label[appendix]{subsec:analytical_1d_gaussian}

The PDF and CDF of a 1-D Gaussian distribution are
\begin{equation}
  \density(\rvar; \parvec)
  =
  \frac{1}{\sigma\sqrt{2\pi}}\exp{\left(\frac{-(\rvar - \mu)^2}{2\sigma^2}\right)}
\,; \qquad
  \cdf(\rvar; \parvec)
  =
  \frac{1}{2}\left[1 + \erf{\left(\frac{\rvar - \mu}{\sigma\sqrt{2}}\right)}\right]
  \,,
\end{equation}
where $\parvec \coloneqq \begin{bsmallmatrix} \mu & \sigma \end{bsmallmatrix}\tran$ is the parameter vector containing the mean and the standard deviation.
The inverse function of the CDF is
\begin{equation}
  x = \cdf\inv(u; \parvec) = \mu + \sigma\sqrt{2}\erf\inv{(2u - 1)}
  \,,
\end{equation}
in which $u=\cdf(x; \parvec)$ is the CDF value at the given $x$ and $\parvec$.

The inverse CDF gives us direct access to the space-parameter sensitivity, leading to \eqref{eqn:1d_gauss_grad}:
\begin{equation}
  \nabla_{\parvec} \rvar = \nabla_{\parvec} \cdf\inv
  =
  \begin{bmatrix} \pderiv{\cdf\inv}{\mu} \\ \pderiv{\cdf\inv}{\sigma} \end{bmatrix}
  =
  \begin{bmatrix}
    1 \\
    \sqrt{2}\erf\inv{(2u - 1)}
  \end{bmatrix}
  =
  \begin{bmatrix}
    1 \\
    \frac{x - \mu}{\sigma}
  \end{bmatrix}
  \,.
\end{equation}

Alternatively, we can compute the sensitivity using \cref{eqn:1d_sensitivity_def}:
\begin{equation}
  \nabla_{\parvec} \rvar
  =
  - \frac{1}{\density}
  \begin{bmatrix} \pderiv{\cdf}{\mu} \\ \pderiv{\cdf}{\sigma} \end{bmatrix}
  =
  - \frac{1}{\density}
  \begin{bmatrix} - \density \\ - \frac{\rvar - \mu}{\sigma} \density \end{bmatrix}
  =
  \begin{bmatrix} 1 \\ \frac{\rvar - \mu}{\sigma} \end{bmatrix}
  \,,
\end{equation}
which also leads to \eqref{eqn:1d_gauss_grad}. 

\subsection{2-D Gaussian Distribution}
\label[appendix]{subsec:analytical_2d_gaussian}

The joint PDF of a 2-D Gaussian distribution at a given space coordinate
$\rvec=\begin{bsmallmatrix} \rvar_1 & \rvar_2 \end{bsmallmatrix}$ is
\begin{equation}
  \density(\rvec; \parvec)
  =
  \dfrac{1}{2 \pi \sigma_1 \sigma_2 \sqrt{1 - \rho^2}}
  \exp\left(
      - \frac{z_1^2 - 2\rho z_1 z_2 + z_2^2}{2\left(1-\rho^2\right)}
  \right) \,.
\end{equation}
Here, $z_1 \coloneqq (\rvar_1-\mu_1)\fracslash\sigma_1$ and $z_2 \coloneqq (\rvar_2-\mu_2)\fracslash\sigma_2$ are defined for simplicity in this subsection.
The parameter vector $\parvec$ is defined as $\{\mu_1, \mu_2, \sigma_1, \sigma_2, \rho\}$, where these correspond to the means and standard deviations of the two dimensions as well as the correlation factor.
The conditional distributions of a 2-D Gaussian are 1-D Gaussians:
\begin{equation}
  \begin{aligned}
    \density_1(\rvar_1 \condon \rvar_2; \parvec)
    &=
    \dfrac{1}{\sigma_{\rvar_1\,|\,\rvar_2} \sqrt{2 \pi}}
    \exp\left(
      -\dfrac{1}{2}\dfrac{\left(\rvar_1-\mu_{\rvar_1\,|\,\rvar_2}\right)^2
        }{\sigma_{\rvar_1\,|\,\rvar_2}^2}
    \right) \,,
    \\
    \density_2(\rvar_2 \condon \rvar_1; \parvec)
    &=
    \dfrac{1}{\sigma_{\rvar_2\,|\,\rvar_1} \sqrt{2 \pi}}
    \exp\left(
      -\dfrac{1}{2}\dfrac{\left(\rvar_2-\mu_{\rvar_2\,|\,\rvar_1}\right)^2
        }{\sigma_{\rvar_2\,|\,\rvar_1}^2}
    \right) \,,
  \end{aligned}
\end{equation}
where the conditional means and standard deviations are
\begin{equation}
  \left\{
    \begin{array}{l}
      \mu_{\rvar_1\,|\,\rvar_2} \coloneqq
        \mu_1 + \rho \sigma_1 z_2 \\
      \sigma_{\rvar_1\,|\,\rvar_2} \coloneqq \sigma_1 \sqrt{1-\rho^2} \\
    \end{array}
  \right. \,,
  \text{ and }
  \left\{
    \begin{array}{l}
      \mu_{\rvar_2\,|\,\rvar_1} \coloneqq
        \mu_2 + \rho \sigma_2 z_1 \\
      \sigma_{\rvar_2\,|\,\rvar_1} \coloneqq \sigma_2 \sqrt{1-\rho^2}
    \end{array}
  \right.
  \,.
\end{equation}
The conditional CDFs thus follow those of 1-D Gaussian CDFs:
\begin{equation}
  \begin{aligned}
    \cdf_1(\rvar_1 \condon \rvar_2; \parvec)
    &=
    \Phi\left(\dfrac{\rvar_1-\mu_{\rvar_1\,|\,\rvar_2}
      }{\sigma_{\rvar_1\,|\,\rvar_2}}\right)
    =
    \dfrac{1}{2}\left[ 1 + \erf\left(
        \dfrac{\rvar_1-\mu_{\rvar_1\,|\,\rvar_2}}{\sigma_{\rvar_1\,|\,\rvar_2}\sqrt{2}}
    \right) \right] \,,
    \\
    \cdf_2(\rvar_2 \condon \rvar_1; \parvec)
    &=
    \Phi\left(\dfrac{\rvar_2 - \mu_{\rvar_2\,|\,\rvar_1}
      }{\sigma_{\rvar_2\,|\,\rvar_1}}\right)
    =
    \dfrac{1}{2}\left[ 1 + \erf\left(
        \dfrac{\rvar_2-\mu_{\rvar_2\,|\,\rvar_1}}{\sigma_{\rvar_2\,|\,\rvar_1}\sqrt{2}}
    \right) \right] \,.
  \end{aligned}
\end{equation}

The gradients of conditional CDFs w.r.t.\ $\parvec$ are
\begin{equation}
  \nabla_{\parvec}
  \begin{bmatrix} \cdf_1 \\ \cdf_2 \end{bmatrix}
  =
  \begin{bmatrix} \density_1 & 0 \\ 0 & \density_2 \end{bmatrix}
  \begin{bmatrix}
    -1 & \rho\frac{\sigma_1}{\sigma_2} &
    - z_1 & \rho\frac{\sigma_1}{\sigma_2}z_2 &
    \frac{\sigma_1}{1-\rho^2}\left[ \rho z_1 - z_2 \right]
    \\
    \rho\frac{\sigma_2}{\sigma_1} & -1 &
    \rho\frac{\sigma_2}{\sigma_1} z_1 & - z_2 &
    - \frac{\sigma_2}{1-\rho^2}\left[ z_1 - \rho z_2 \right]
  \end{bmatrix} \,,
\end{equation}
The gradients of conditional CDFs w.r.t.\ $\rvec$ are
\begin{equation}
  \nabla_{\rvec}
  \begin{bmatrix}
    \cdf_1(\cdot) \\
    \cdf_2(\cdot)
  \end{bmatrix}
  =
  \begin{bmatrix}
    \density_1 & - \rho \frac{\sigma_1}{\sigma_2} \density_1 \\
    - \rho \frac{\sigma_2}{\sigma_1} \density_2 & \density_2
  \end{bmatrix} \,.
\end{equation}

Applying $\nabla_{\parvec} \rvec = - \left[\nabla_{\rvec} \vec{\cdf}\right]\inv \nabla_{\parvec} \vec{\cdf}$, we obtain \eqref{eqn:2d_gauss_grad}
\begin{equation}\label{eqn:2D:Gauss:Transform}
\begin{aligned}
  \nabla_{\parvec} \begin{bmatrix} \rvar_1 \\ \rvar_2 \end{bmatrix}
  &=
  -
  \begin{bmatrix}
      \density_1 & - \rho \frac{\sigma_1}{\sigma_2} \density_1 \\
      - \rho \frac{\sigma_2}{\sigma_1} \density_2 & \density_2
  \end{bmatrix}^{-1}
  \left(
    \nabla_{\parvec} \begin{bmatrix} \cdf_1 \\ \cdf_2 \end{bmatrix} 
  \right)
  \\
  &=
  -
  \begin{bmatrix}
      \frac{1}{\density_1}\frac{1}{1-\rho^2}
        & \rho \frac{\sigma_1}{\sigma_2}\frac{1}{\density_2}\frac{1}{1-\rho^2}
      \\
      \rho \frac{\sigma_2}{\sigma_1}\frac{1}{\density_1}\frac{1}{1-\rho^2}
        & \frac{1}{\density_2}\frac{1}{1-\rho^2}
  \end{bmatrix}
  \left(
  \nabla_{\parvec} \begin{bmatrix} \cdf_1 \\ \cdf_2 \end{bmatrix}
  \right)
  =
  \begin{bmatrix}
      1 & 0 & z_1 & 0 & \frac{\sigma_1}{1-\rho^2} z_2 
      \\
      0 & 1 & 0 & z_2 & \frac{\sigma_2}{1-\rho^2} z_1
  \end{bmatrix}\,.
\end{aligned}
\end{equation}

To further approximate the Jacobian with the diagonal terms as described in \Cref{subsec:inverse_mat_approx}, we have
\begin{equation}\begin{aligned}
  \nabla_{\parvec} \begin{bmatrix} \rvar_1 \\ \rvar_2 \end{bmatrix}
  &=
  -
  \begin{bmatrix} \frac{1}{\density_1} & 0 \\ 0 & \frac{1}{\density_2} \end{bmatrix}
  \left(
  \nabla_{\parvec} \begin{bmatrix} \cdf_1 \\ \cdf_2 \end{bmatrix}
  \right)
  \\
  &=
  \begin{bmatrix}
    1 & -\rho\frac{\sigma_1}{\sigma_2} &
    z_1 & -\rho\frac{\sigma_1}{\sigma_2} z_2 &
    -\frac{\sigma_1}{1-\rho^2}\left( \rho z_1 - z_2 \right)
    \\
    -\rho\frac{\sigma_2}{\sigma_1} & 1 &
    -\rho\frac{\sigma_2}{\sigma_1} z_1 & z_2 &
    \frac{\sigma_2}{1-\rho^2}\left( z_1 - \rho z_2 \right)
  \end{bmatrix}
  \,,
\end{aligned}
\end{equation}
which gives \eqref{eqn:2d_gauss_grad_approx}.

  \iftrue
  \begin{center}
    \scriptsize
    \framebox{
      \parbox{\linewidth}{%
      Government License (will be removed at publication): %
      The submitted manuscript has been created by UChicago Argonne, LLC, Operator of Argonne National Laboratory (``Argonne").  Argonne, a U.S. Department of Energy Office of Science laboratory, is operated under Contract No. DE-AC02-06CH11357. %
      The U.S. Government retains for itself, and others acting on its behalf, a paid-up nonexclusive, irrevocable worldwide license in said article to reproduce, prepare derivative works, distribute copies to the public, and perform publicly and display publicly, by or on behalf of the Government. %
      The Department of Energy will provide public access to these results of federally sponsored research in accordance with the DOE Public Access Plan. %
      http://energy.gov/downloads/doe-public-access-plan.%
      }
    }
    \normalsize
  \end{center}
  \fi
\end{document}